\documentclass[journal]{IEEEtran}
\usepackage{cite}
\ifCLASSINFOpdf
\usepackage[pdftex]{graphicx}
\else
\usepackage[dvips]{graphicx}
\fi
\usepackage{amsmath}

\DeclareMathOperator*{\argmin}{arg\,min}
\usepackage{amssymb}
\usepackage{amsfonts}
\usepackage{amsopn}
\usepackage{array}
\usepackage{booktabs}
\usepackage[table]{xcolor}
\usepackage[caption=false,font=normalsize,labelfont=sf,textfont=sf]{subfig}
\usepackage[linesnumbered,ruled]{algorithm2e}
\usepackage{url}
\usepackage{microtype}
\usepackage{multirow}
\usepackage{textcomp}
\usepackage{xcolor}

\begin{document}
\title{A Survey on Evolutionary Neural Architecture Search}
\author{Yuqiao~Liu,~\IEEEmembership{Student Member,~IEEE,}
	Yanan~Sun,~\IEEEmembership{Member,~IEEE,}
	Bing~Xue,~\IEEEmembership{Member,~IEEE,}
	Mengjie~Zhang,~\IEEEmembership{Fellow,~IEEE,}
	Gary G. Yen,~\IEEEmembership{Fellow,~IEEE}
	~and~Kay Chen Tan,~\IEEEmembership{Fellow,~IEEE}
	\thanks{Yuqiao Liu and Yanan Sun are with the College of Computer Science, Sichuan University, China.}
	\thanks{Bing Xue and Mengjie Zhang are with the School of Engineering and Computer Science, Victoria University of Wellington, Wellington, New Zealand.}
	\thanks{Gary G. Yen is with the School of Electrical and Computer Engineering, Oklahoma State University, Stillwater, OK, USA.}
	\thanks{Kay Chen Tan is with the Department of Computing, Hong Kong Polytechnic University, Hong Kong.}
	}

{}

\maketitle
\begin{abstract}
Deep Neural Networks (DNNs) have achieved great success in many applications. The architectures of DNNs play a crucial role in their performance, which is usually manually designed with rich expertise. However, such a design process is labour intensive because of the trial-and-error process, and also not easy to realize due to the rare expertise in practice. Neural Architecture Search (NAS) is a type of technology that can design the architectures automatically. Among different methods to realize NAS, Evolutionary Computation (EC) methods have recently gained much attention and success. Unfortunately, there has not yet been a comprehensive summary of the EC-based NAS algorithms. This paper reviews over 200 papers of most recent EC-based NAS methods in light of the core components, to systematically discuss their design principles as well as justifications on the design. Furthermore, current challenges and issues are also discussed to identify future research in this emerging field.
\end{abstract}

\IEEEpeerreviewmaketitle

\section{Introduction}
\IEEEPARstart{D}{eep} Neural Networks (DNNs), as the cornerstone of deep learning~\cite{lecun2015deep}, have demonstrated their great success in diverse real-world applications, including image classification~\cite{he2016deep,huang2017densely}, natural language processing~\cite{devlin2018bert}, speech recognition~\cite{zhang2017very}, to name a few. The promising performance of DNNs has been widely documented due to their deep architectures~\cite{lecun2015deep}, which can learn meaningful features directly from the raw data almost without any explicit feature engineering. Generally, the performance of DNNs depends on two aspects: their architectures and the associated weights. Only when both achieve the optimal status simultaneously, the performance of the DNNs could be expected promising. The optimal weights are often obtained through a learning process: using a continuous loss function to measure the discrepencies between the real output and the desired one, and then the gradient-based algorithms are often used to minimize the loss. When the termination condition is satisfied, which is commonly a maximal iteration number, the algorithm can often find a good set of weights~\cite{lecun2015deep}. This kind of process has been very popular owing to its effectiveness in practice, and has become the dominant practice for weight optimization~\cite{bengio2012practical}, although they are principally local-search~\cite{kearney1987optical} algorithms. On the other hand, obtaining the optimal architectures cannot be directly formulated by a continuous function, and there is even no explicit function to measure the process for finding optimal architectures.

To this end, there has been a long time that the promising architectures of DNNs are manually designed with rich expertise. This can be evidenced from the state of the arts, such as VGG~\cite{simonyan2014very}, ResNet~\cite{he2016deep} and DenseNet~\cite{huang2017densely}. These promising Convolutional Neural Network (CNN) models are all manually designed by the researchers with rich knowledge in both neural networks and image processing. However, in practice, most end users are not with such kinds of knowledge. Moreover, DNN architectures are often problem-dependent. If the distribution of the data is changed, the architectures must be redesigned accordingly. Neural Architecture Search (NAS), which aims to automate the architecture designs of deep neural networks, is identified as a promising way to address the challenge aforementioned. 

Mathematically, NAS can be modeled by an optimization problem formulated by Equation~(\ref{equ_defination_NAS}):
\begin{equation}
\label{equ_defination_NAS}
\left\{
\centering
\begin{array}{c}
\argmin_A = \mathcal{L}(A, \mathcal{D}_{train}, \mathcal{D}_{fitness}) \\
s.t.\quad  A \in \mathcal{A} \\
\end{array}
\right.
\end{equation}
where $\mathcal{A}$ denotes the search space of the potential neural architectures, $\mathcal{L}(\cdot)$ measures the performance of the architecture $A$ on the fitness evaluation dataset $\mathcal{D}_{fitness}$ after being trained on the training dataset $\mathcal{D}_{train}$. The $\mathcal{L}(\cdot)$ is usually non-convex and non-differentiable~\cite{sun2019completely}. In principle, NAS is a complex optimization problem experiencing several challenges, e.g., complex constraints, discrete representations, bi-level structures, computationally expensive characteristics and multiple conflicting criteria. NAS algorithms refer to the optimization algorithms which are specifically designed to effectively and efficiently solve the problem represented by Equation~(\ref{equ_defination_NAS}). The initial work of NAS algorithms is commonly viewed as the work in~\cite{zoph2016neural}, which was proposed by Google. The pre-print version of this work was firstly released in the website of arXiv\footnote{\url{https://arxiv.org/abs/1611.01578}} in 2016, and then was formally accepted for publication by the International Conference on Learning Representations (ICLR) in 2017. Since then, a large number of researchers have been investing tremendous efforts in developing novel NAS algorithms.

Based on the optimizer employed, existing NAS algorithms can be broadly classified into three different categories: Reinforcement Learning (RL)~\cite{kaelbling1996reinforcement} based NAS algorithms, gradient-based NAS algorithms, and Evolutionary Computation (EC)~\cite{back1997handbook} based NAS algorithms (ENAS). Specifically, RL based algorithms often require thousands of Graphics Processing Cards (GPUs) performing several days even on median-scale dataset, such as the CIFAR-10 image classification benchmark dataset~\cite{krizhevsky2009learning}. Gradient-based algorithms are more efficient than RL based algorithms. However, they often find the ill-conditioned architectures due to the improper relation for adapting to gradient-based optimization. Unfortunately, the relation has not been mathematically proven. In addition, the gradient-based algorithms require to construct a supernet in advance, which also highly requires expertise. The ENAS algorithms solve the NAS by exploiting EC techniques. Specifically, EC is a class of population-based computational paradigms, simulating the evolution of species or the behaviors of the population in nature, to solve challenging optimization problems. In particular, Genetic Algorithms (GAs)~\cite{goldberg2006genetic}, Genetic Programming (GP)~\cite{banzhaf1998genetic}, and Particle Swarm Optimization (PSO)~\cite{kennedy1995particle} are widely used EC methods in practice. Owing to the promising characteristics of EC methods in insensitiveness to the local minima and no requirement to gradient information, EC has been widely applied to solve complex non-convex optimization problems~\cite{sun2018igd}, even when the mathematical form of the objective function is not available~\cite{darwish2020survey}.

\begin{figure}
	\centering
	\includegraphics[width=0.8\linewidth]{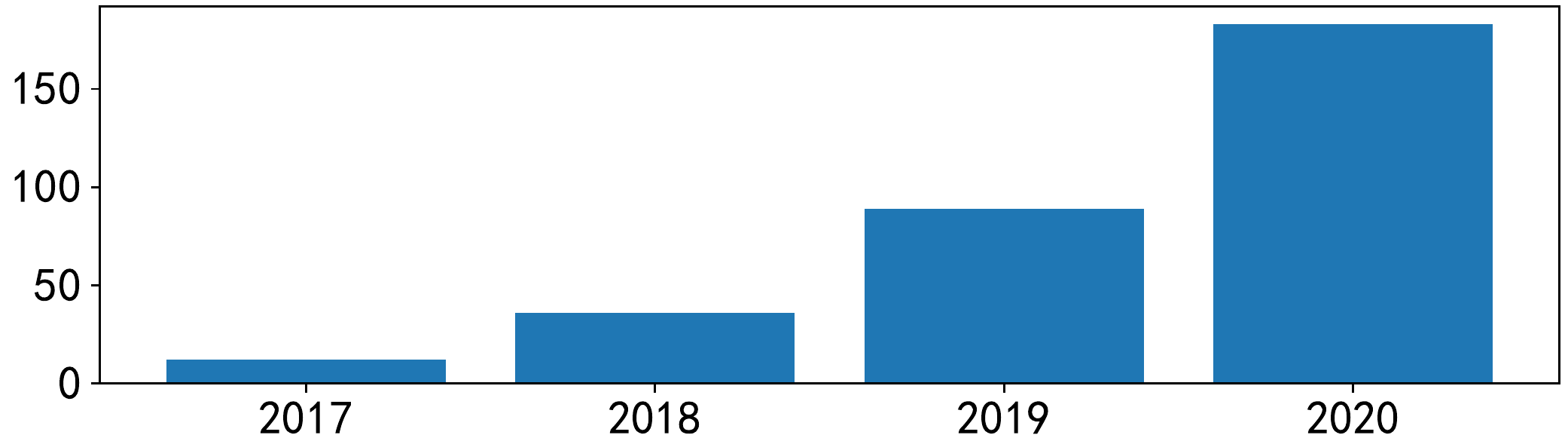}
	\caption{The number of submissions refers to the works of evolutionary neural architecture search. The data is collected from Google Scholar with the keywords of ``evolutionary" OR ``genetic algorithm'' OR ``particle swarm optimization" OR ``PSO" OR ``genetic programming" AND ``architecture search" OR ``architecture design'' OR ``CNN'' OR ``deep learning'' OR ``deep neural network'' and the literature on Neural Architecture Search collected from the AutoML.org website by the end of 2020. With these initially collected data, we then have carefully checked each manuscript to make its scope accurately within the evolutionary neural architecture search.}
	\label{fig_histogram}
	\vspace{-0.3cm}
\end{figure}

In fact, the EC methods had been frequently used more than twenty years ago, searching for not only the optimal neural architectures but also the weights of neural networks simultaneously, which is also termed as neuroevolution~\cite{floreano2008neuroevolution}. The major differences between ENAS and neuroevolution lie in two aspects. Firstly, neuroevolution often uses EC to search for both the neural architectures and the optimal weight values, while ENAS for now focuses mainly on searching for the architectures and the optimal weight values are obtained by using the gradient-based algorithms\footnote{Please note that we still categorize some existing algorithms as the ENAS algorithm, such as API~\cite{dufourq2017automated}, EvoCNN~\cite{sun2019evolving} and EvoDeep~\cite{martin2018evodeep}, although they also concern the weights. This is because the optimal weight values of the DNNs searched by them are still obtained by the gradient-based algorithms. They only searched for the best weight initialization values or the best weight initialization method of the DNNs.} immediately after. Secondly, neuroevolution commonly applies to small-scale and median-scale neuron networks, while ENAS generally works on DNNs, such as the deep CNNs~\cite{real2017large, sun2020automatically} and deep stacked autoencoders~\cite{sun2018particle}, which are stacked by the building blocks of deep learning techniques~\cite{lecun2015deep}. Generally, the first work of ENAS is often viewed as the LargeEvo algorithm~\cite{real2017large} which was proposed by Google who released its early version in March 2017 in arXiv. Afterwards, this paper got accepted into the 34th International Conference on Machine Learning in June 2017. The LargeEvo algorithm employed a GA to search for the best architecture of a CNN, and the experimental results on CIFAR-10 and CIFAR-100~\cite{krizhevsky2009learning} have demonstrated its effectiveness. Since then, a large number of ENAS algorithms have been proposed. Fig.~\ref{fig_histogram} shows the number of similar works\footnote{These ``submissions" include the ones which have been accepted for publication after the peer-review process, and also the ones which are only available on the arXiv website without the peer-review process.} published from 2017 to 2020 when this survey paper was ready for submission. As can be seen from Fig.~\ref{fig_histogram}, from 2017 to 2020, the number of submissions grows with multiple scales.

A large number of related submissions have been made available publicly, but there is no comprehensive survey of the literature on ENAS algorithms. Although recent reviews on NAS have been made in~\cite{elsken2018neural,wistuba2019survey,darwish2020survey,stanley2019designing}, they mainly focus on reviewing different methods to realize NAS, rather than concentrating on the ENAS algorithms. Specifically, Elsken \textit{et al.}~\cite{elsken2018neural} divided NAS into three stages: search space, search strategy, and performance estimation strategy. Similarly, Wistuba \textit{et al.}~\cite{wistuba2019survey} followed these three stages with an additional review about the multiple objectives in NAS. Darwish \textit{et al.}~\cite{darwish2020survey} made a summary of Swarm Intelligence (SI) and Evolutionary Algorithms (EAs) approaches for deep learning, with the focuses on both NAS and other hyperparameter optimization. Stanley \textit{et al.} ~\cite{stanley2019designing} went through a review of neuroevolution, aiming at revealing the weight optimization rather than the architecture of neural networks. Besides, most of the references in these surveys are pre-2019 and do not include an update on the papers published during the past two years when most ENAS works were published. This paper presents a survey involving a large number of ENAS papers, with the expectation to inspire some new ideas for enhancing the development of ENAS. To allow readers easily concentrating on the technical part of this survey, we also follow the three stages to introduce ENAS algorithms, which has been widely adopted by existing NAS survey papers~\cite{elsken2018neural,wistuba2019survey,darwish2020survey}, but with essential modifications made to specifically suit ENAS algorithms.

The remainder of this paper is organized as follows. The background of ENAS is discussed in Section~\ref{sec_background}. Section~\ref{sec_encoding_space} documents different encoding spaces, initial spaces and search spaces of ENAS algorithms. In Section~\ref{sec_architectureEncoding}, the encoding strategy and architecture representation are introduced. Section~\ref{sec_Population_operators} summarizes the process of population updating, including evolutionary operators and selection strategies. Section~\ref{sec_evaluation} shows multiple ways to speed up the evolution. Section~\ref{sec_application} presents the applications of ENAS algorithms. Section~\ref{sec_challenges_and_issues} discusses the challenges and prospects, and finally Section~\ref{sec_conclusion} presents conclusions.

\section{Background}
\label{sec_background}
As discussed above, ENAS is a process of searching for the optimal architectures of DNNs by using EC methods. In this section, we will first introduce the unified flow chart of ENAS in Subsection~\ref{sec2_3}. Then, the categories of EC methods based on their search strategies and the categories of DNN architectures in Subsections~\ref{sec2_1} and \ref{sec2_2}, respectively. 

\subsection{Flow Chart of ENAS}
\label{sec2_3}

\begin{figure}
	\centering
	\includegraphics[width=0.9\linewidth]{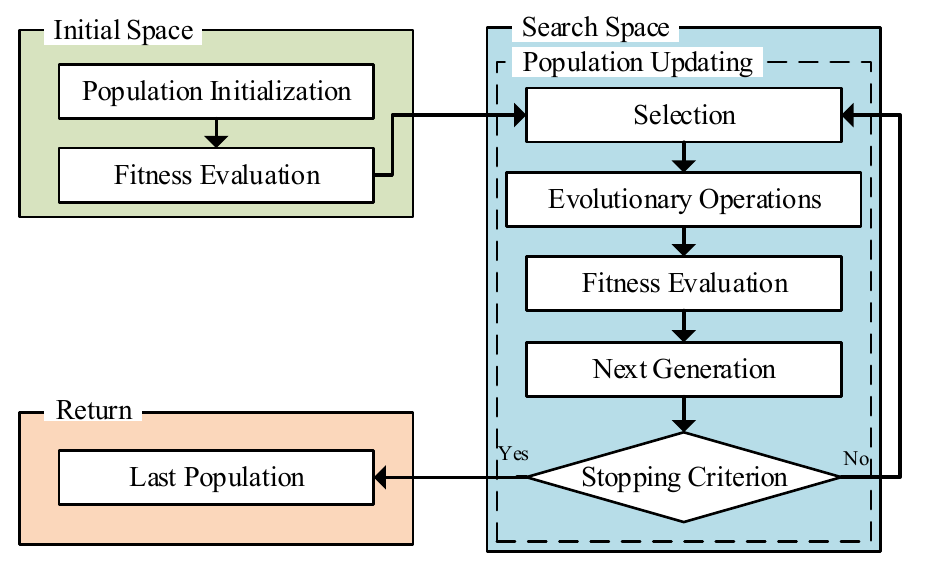}
	\caption{The flowchart of a common ENAS algorithm.}
	\label{fig_flowchart}
	\vspace{-0.3cm}
\end{figure}
Fig.~\ref{fig_flowchart} shows an illustration of the flowchart of an ENAS algorithm. Specifically, the evolutionary process takes place in the initial space and the search space sequentially. First of all, a population is initialized within the initial space that is defined in advance. Each individual in the population represents a solution for the ENAS, i.e., a DNN architecture. Each architecture needs to be encoded as an individual before it joins the population. Second, the fitness of the generated individuals is evaluated. Note that there are two fitness evaluation steps as shown in Fig.~\ref{fig_flowchart}, which commonly employ the same evaluation criterion. Thirdly, after the fitness evaluation of the initial population, the whole population starts the evolutionary process within the search space, which is shown by the dashed box in Fig.~\ref{fig_flowchart}. In the evolutionary process, the population is updated by the selection and the evolutionary operators in each iteration, until the stopping criterion is met. Please note that the selection stage is not necessary for some other EC paradigms like SI. Finally, a population that has finished the evolution is obtained. In the following sections, these key steps are documented in detail.

\subsection{Evolutionary Search Strategy}
\label{sec2_1}
\begin{figure}
	\centering
	\includegraphics[width=1\linewidth]{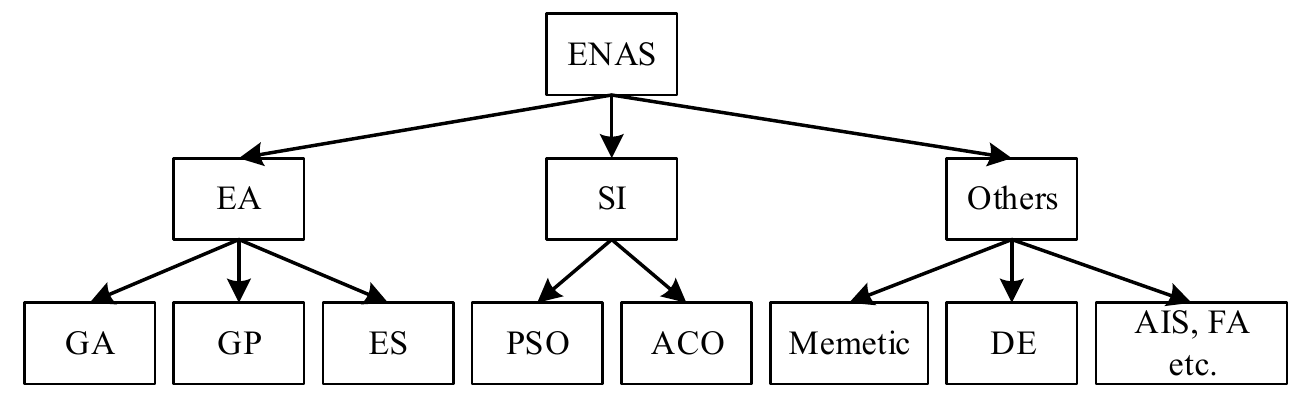}
	\caption{The categories of ENAS from EC methods regarding the search strategies.}
	\label{fig_EC_taxonomy}
	\vspace{-0.3cm}
\end{figure}

ENAS is distinguished from other NAS methods by its employed optimization approach, i.e., the EC methods, where the optimization approaches can be further subdivided based on the search strategy that the optimization approach adopted. Fig.~\ref{fig_EC_taxonomy} provides such an illustration from three aspects: EAs, Swarm Intelligence (SI), and others, and more detailed statistics can be observed in Table~\ref{table_EC}. In practice, the EA-based methods account for the majority of existing ENAS algorithms, where the GA takes a large part of EA-based methods. The other categories of EC methods are also important parts for realizing ENAS algorithms, such as GP, Evolutionary Strategy (ES), PSO, Ant Colony Optimization (ACO)~\cite{dorigo1996ant}, Differential Evolution (DE)~\cite{price2006differential}, Firefly Algorithm (FA)~\cite{yang2009firefly}. In this paper, the Hill-Climbing Algorithm (HCA) is classified into EC paradigm because it can be regarded as an EA with a simple selection mechanism and without crossover operation~\cite{elsken2017simple}. HCA has been well known as a widely used local search algorithm in memetic algorithms~\cite{wang2009memetic}.

\subsection{Common Neural Networks in ENAS}
\label{sec2_2}

\begin{figure}
	\centering
	\includegraphics[width=0.9\linewidth]{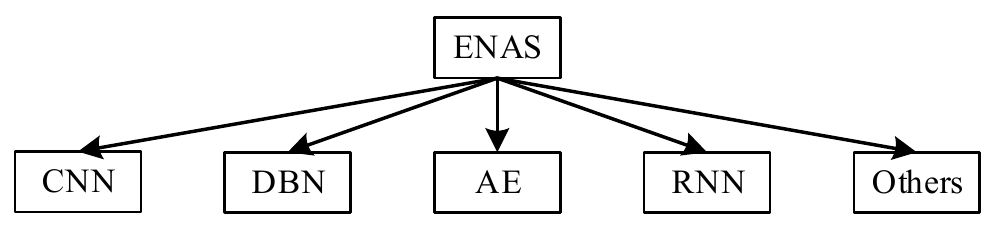}
	\caption{The categories of ENAS from neural network perspective.}
	\label{fig_NN_taxonomy}
	\vspace{-0.3cm}
\end{figure}

\begin{figure}
	\centering
	\includegraphics[width=1\linewidth]{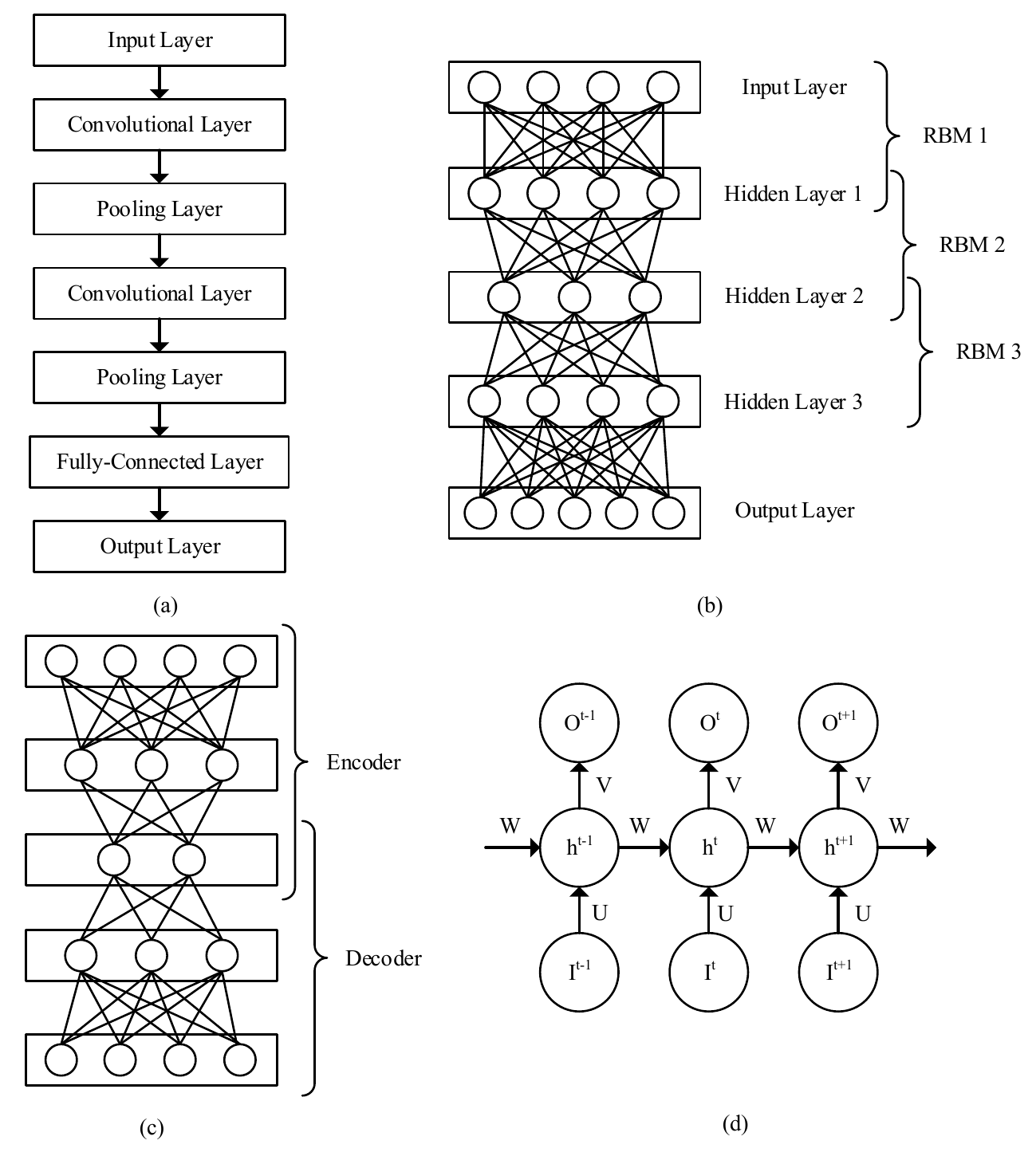}
	\caption{Examples of different neural architectures in ENAS. (a) CNN. (b) DBN. (c) AE. (d) RNN. }
	\label{fig_NN}
	\vspace{-0.3cm}
\end{figure}

The end product of ENAS is all about DNNs, which can be broadly divided into five different categories: CNN, Deep Belief Network (DBN), Stacked Auto-Encoder (SAE), Recurrent Neural Network (RNN) and others. The brief fact can be seen from Fig.~\ref{fig_NN_taxonomy}, and a detailed illustration will be shown in Table~\ref{table_EC}. 

Most ENAS methods are proposed for searching for the optimal CNN architectures. This is because many hand-crafted CNNs, such as VGG~\cite{simonyan2014very}, ResNet~\cite{he2016deep} and DenseNet~\cite{huang2017densely}, have demonstrated their superiority in handling the image classification tasks which is the most successful applications in the deep learning field. Generally, the CNN architecture is composed of the convolutional layers, the pooling layers and the fully-connected layers, and a common example of CNNs is shown in Fig.~\ref{fig_NN} (a). The optimization of CNN architecture is mainly composed of three aspects: the hyperparameters of each layer~\cite{fujino2017deep}, the depth of the architecture~\cite{sun2019evolving} and the connections between layers~\cite{xie2017genetic}. In practice, the majority of the ENAS methods consider the above three aspects collectively~\cite{real2017large, real2019regularized}.

DBN~\cite{hinton2006fast} is made up by stacking multiple Restricted Boltzmann Machines (RBMs), and an example can be seen in Fig.~\ref{fig_NN} (b). Specifically, RBMs have connections only between layers, while without any inter-layer connection. Meanwhile, RBMs allow the DBN to be trained in an unsupervised manner to obtain a good weight initialization. There are two commonly used types of hyperparameters needed to be optimized in DBN: the number of neurons in each layer~\cite{kim2017particle} and the number of layers~\cite{qiang2019neural, zhang2019identify}.

An SAE is constructed by stacking multiple of its building blocks which are AEs. An AE aims to learn meaningful representations from the raw input data by restoring its input data as its objective~\cite{goodfellow2016deep}. Generally, an AE is typically composed of two symmetrical components: the encoder and the decoder, and an example including both parts is shown in Fig.~\ref{fig_NN} (c). Generally, some ENAS methods only take the encoder into evolution~\cite{suganuma2018exploiting, sun2018particle} because the decoder part is symmetric with the encoder and can be derived accordingly. Yet, some other ENAS methods optimize the hyperparameters of encoders and decoders separately~\cite{hajewski2020evolutionary, rodriguez2019evolving}.

The most significant difference between RNNs and the neural networks introduced above is its recurrent connection. Fig.~\ref{fig_NN} (d) shows the time-expanded structure of an RNN, where the value of current hidden layer $h^{t}$ is influenced by the value at its previous time slot $h^{t-1}$ and the output value of its previous layer. Because these layers are reused, all the weights (i.e., \textit{U}, \textit{W} and \textit{V} in the figure) are shared. Different from focusing on the number of neurons and the number of layers in the feedforward neural network, some ENAS methods concern about how many times the RNN should be unfold~\cite{camero2018evolutionary, camero2019specialized}. In addition, there remain other neural networks like typical DNNs, which are made up of only fully-connected layers, where the connections are formed by all the neurons in the two adjacent layers. Because such kind of DNNs is not the major target investigated by NAS, we will not introduce it in detail.

When the ENAS algorithms are applied to these DNNs, the goal is to find the best architecture-related parameters. Specifically, for CNNs, they are the number of convolutional layers, pooling layers, fully-connected layers, and parameters related to these layers (such as the kernel size of the convolutional layers, the pooling type, and the number of neurons for fully-connected layers, and so on) as well as their connection situation (such as the dense connection and the skip connection). For DBN and SAE, they are the number of their building blocks, i.e., the RBM for DBN and the AE for SAE, and the number of neurons in each layer. For RNN, in addition to the architecture-related parameters mentioned above, the number of the time slot is also an important parameter to be optimized by ENAS algorithms. For the traditional DNNs, the ENAS algorithms often concern about the neuron number of each layer. In addition, some ENAS algorithms also concern the weights, such as the weight initialization method and weigh initial values. Table~\ref{table_parameters} summarizes the detail of the common parameters optimized by ENAS methods in different types of neural networks, including the CNNs as the most popular type. The details of these ENAS algorithms will be documented in the following sections. 

\begin{table}[]
	\renewcommand\arraystretch{0.8}
	\caption{Common Parameters Optimized in Different DNNs.} 
	\label{table_parameters}
	\centering
	\begin{tabular}{p{1.2cm}|p{1.5cm}|p{4.5cm}}
		\hline
		& \multicolumn{2}{c}{Parameters}                                                                                                                                                                             \\ \hline
		\multirow{4}{*}{CNN} & global parameters       & number of layers, connections between layers                                                                                                               \\ \cline{2-3} 
		& convolution layer       & filter size (width and height), stride size (width and height), feature map size, convolution type, standard deviation and mean value of the filter elements \\ \cline{2-3} 
		& pooling layer           & filter size (width and height), stride size (width and height), pooling type                                                                              \\ \cline{2-3} 
		& fully-connected layer   & number of neurons, standard deviation and mean value of weights                                                                                             \\ \hline
		DBN, AE              & \multicolumn{2}{l}{number of hidden layers, neurons per layer}                                                                                                                                             \\ \hline
		RNN                  & \multicolumn{2}{l}{\shortstack[l]{number of hidden layers, neurons per layer, number of\\ time slot}}                                                                                                                        \\ \hline
	\end{tabular}
\vspace{-0.3cm}
\end{table}

\section{Encoding Space}
\label{sec_encoding_space}

\begin{table*}[]
	\renewcommand\arraystretch{0.8}
	\caption{Different Encoding Space and the Constraints} 
	\label{table_constrains}
	\vspace{-0.3cm}
	\begin{center}
		\begin{tabular}{p{3.8cm}|p{1.8cm}|p{2cm}|p{2.5cm}|p{5.5cm}}
			\hline
			& Fixed depth & Rich initialization & Partial fixed structure & Relatively few constraints\\
			\hline
			Layer-based encoding space & ~\cite{sun2018experimental, singh2019genetic, dahou2019arabic, shu2020automatically, li2019day, fujino2017deep, saufi2018differential, kang2019efficient, cheung2011hybrid, zhang2016multiobjective, ye2017particle, fujino2018recognizing} & ~\cite{tanaka2016automated, fujino2017deep, kwasigroch2019deep, zhu2019eena, johner2019efficient, shen2019searching} & ~\cite{bi2019evolutionary,  gibb2018genetic,  zhang2019new, ahmed2019novel,  tian2019automated, dufourq2017automated, wei2019automatic, ortego2020evolutionary, piergiovanni2019evolving, cheung2011hybrid, prellberg2018lamarckian, young2015optimizing, calisto2020self} & ~\cite{real2017large, junior2019particle, wang2018hybrid, frachon2019immunecs, qiang2019neural, sun2018particle, camero2019specialized, sharaf2020automated, sabar2019evolutionary, sapra2020evolutionary, sun2019evolving, kwasigroch2020neural, schorn2019automated,  assunccao2019automatic, laredo2019automatic, anwar2019boosting, teng2019catalytic, litzinger2019compute, sapra2020constrained, martinez2020coronavirus, rapaport2019eegnas, peng2018effective, dahal2020effective, jiang2020efficient, ren2019eigen, chung2019emotion, martin2018evodeep, almalaq2018evolutionary, van2019evolutionary, wang2018evolving, wang2019evolving, suganuma2018exploiting, assunccao2019fast, liu2017hierarchical, zhang2019identify, lu2019multi, lorenzo2017particle, passricha2019pso, zhang2020sampled, elsken2018efficient, byla2019deepswarm}\\
			\hline
			Block-based encoding space & ~\cite{baldeon2019adaresu, chen2019auto, wang2019evolving1} & ~\cite{fielding2018evolving} & ~\cite{ortego2020evolutionary, fielding2018evolving, cetto2019size} & ~\cite{elsken2017simple, frachon2019immunecs, suganuma2017genetic,  sun2020automatically, sun2019completely, loni2018designing, song2019efficient, loni2020deepmaker, suganuma2020evolution, miikkulainen2019evolving, hassanzadeh2020evou, miahi2019genetic, mitschke2018gradient, chu2019multi, chen2018reinforced, elsken2018efficient} \\
			\hline
			Cell-based encoding space & & & ~\cite{chen2019efficient, fan2020evolutionary, chen2020immunetnas} & ~\cite{liu2019deep, chen2018reinforced, wistuba2018deep, kramer2018evolution, chu2019fast, chen2018joint, wang2019particle, real2019regularized, saltori2019regularized}\\
			\hline
			Topology-based encoding space &  & ~\cite{wu2019multi, hu2018novel} & & ~\cite{xie2017genetic, lorenzo2018memetic, elsaid2018optimizing, yang2019cars, desell2015evolving, guo2019single, shinozaki2015structure} \\
			\hline
		\end{tabular}
	\end{center}
	\vspace{-0.5cm}
\end{table*}

The encoding space contains all the valid individuals encoded in the population. In terms of the basic units, the encoding space can be divided into three categories according to the basic units they adopt. They are the layer-based encoding space, the block-based encoding space and the cell-based encoding space. In addition, some ENAS methods do not take the configuration of the basic unit into consideration, but care the connections between units. Such kind of encoding space is often called the topology-based encoding space.

In addition, the constraints on the encoding space are important. This is because they represent the human intervention which restricts the encoding space and lightens the burden of the evolutionary process. A method with a mass of constraints can obtain a promising architecture easily but prevent to design any novel architecture that does not follow the constraints. Furthermore, the different sizes of the search space would greatly affect the efficiency of the evolution. On the other hand, the effectiveness of the ENAS methods cannot be guaranteed if there is no constraint on the search space: one extreme case is that all the individuals in the search space are mediocre. In this case, an excellent individual can be obtained even without selection. Table~\ref{table_constrains} shows different kinds of encoding spaces and the constraints of existing ENAS algorithms, where the first row shows the constraints and the first column displays the encoding space. In the following, we will discuss them at length.

\subsection{Encoding Space and Constraints}

The layer-based encoding space denotes that the basic units in the encoding space are the primitive layers, such as convolution layers and fully-connected layers. This would lead to a huge search space, since it tries to encode so much information in the search space. However, it may take more time to search for a promising individual because there are more possibilities to construct a well-performed DNN from the primitive layers. For example, a promising CNN is commonly composed of hundreds of primitive layers, accordingly, the search process will consume more time to search for such a deep architecture by iteratively stacking the primitive layers. In addition, the promising DNN may not be found if only with the primitive layers. For instance, the promising performance of ResNet is widely recognized due to its skip connections which cannot be represented by the primitive layers.

To alleviate the above problems, the block-based encoding space is developed, where various layers of different types are combined as the blocks to serve as the basic unit of the encoding space. Traditional blocks are ResBlock~\cite{he2016deep}, DenseBlock~\cite{huang2017densely}, ConvBlock (Conv2d + BatchNormalization + Activation)~\cite{elsken2017simple} and InceptionBlock~\cite{szegedy2015going}, etc. Specifically, layers in the blocks have a specific topological relationship, such as the residual connection in ResBlock and the dense connections in DenseBlock. These blocks have promising performance and often require fewer parameters to build the architecture. So it is principally easier to find a good architecture in the block-based encoding space compared to the layer-based encoding space. Some ENAS algorithms used these blocks directly, such as~\cite{sun2019completely,frachon2019immunecs}, while other methods proposed different blocks for different purposes. For example, Chen \textit{et al.}~\cite{chen2019auto} proposed eight blocks including ResBlock and InceptionBlock encoded in a 3-bit string, and used Hamming distance to determine the similar blocks. Song \textit{et al.}~\cite{song2019efficient} proposed three residual-dense-mixed blocks to reduce the amount of computation due to the convolution operation of image super-resolution tasks. 

The cell-based encoding space is similar to the block-based one, and can be regarded as a special case of the block-based space where all the blocks are the same. The ENAS algorithms employing this space build the architectures by stacking repeated motifs. Chu \textit{et al.}~\cite{chu2019fast} divided the cell-based space into two independent parts: the micro part containing the parameters of cells, and the macro part defining the connections between different cells. To be more specific, the cell-based encoding space concentrates more on the micro part. Different from the block-based space, the layers in the cell can be combined more freely, and the macro part is always determined by human expertise~\cite{ying2019bench, dong2020bench, real2019regularized}. Some widely used encoding spaces are classified in this category. For example, NAS-Bench-101~\cite{ying2019bench} and NAS-Bench-201~\cite{dong2020bench} search for different combinations of layers and connections in cells, and then stacked the cells sequentially. In addition, NASNet~\cite{zoph2018learning} and DARTS~\cite{liu2018darts} search for two kinds of cells, namely normal cell and reduction cell, and each stacked cell is connected to the two previous cells. The cell-based space greatly reduces the size of the encoding space. This is because all the basic units in the encoding space are the same, and the number of parameters in terms of constructing the promising DNN is much fewer. However, Frachon \textit{et al.}~\cite{frachon2019immunecs} claimed that there is no theoretical basis for that the cell-based space can help to obtain a good architecture. 

In contrast, the topology-based space does not consider the parameters or the structure of each unit (layer or block), yet they are only concerned about the connections between units. One classical example is the one-shot method which treats all the architectures as different subgraphs of a supergraph~\cite{elsken2018neural}. Yang \textit{et al.}~\cite{yang2019cars} proposed CARS to search for the architecture by choosing different connections in the supergraph, and then the architecture was built by the chosen subgraph. CARS can be classified into the topology-based category because it aims at deciding whether or not to keep the connections in the supergraph. But from the perspective of building the architecture, CARS can also be classified into the cell-based category. This is because the subgraph cell was stacked several times to build the architecture. Another typical case is pruning. Wu \textit{et al.}~\cite{wu2019multi} employed a shallow VGGNet~\cite{simonyan2014very} on CIFAR-10, and the aim was to prune unimportant weight connections from the VGGNet.

As observed from Table~\ref{table_constrains}, the constraints on the encoding space mainly focus on three aspects: fixed depth, rich initialization and partial fixed structure. The fixed depth means all the individuals in the population have the same depth. The fixed depth is a strong constraint and largely reduces the size of the encoding space. Please note that the fixed depth is different from the \textit{fixed-length encoding strategy} which will be introduced in Section~\ref{sec_architectureEncoding}. In Genetic CNN~\cite{xie2017genetic}, for example, the \textit{fixed-length encoding strategy} only limits the maximum depth. The node which is isolated (no connection) is simply ignored. By this way, the individuals can obtain different depths. The second constraint is rich initialization (i.e., the \textit{well-designed space} to be discussed in Section~\ref{sec_initial_space}), and it is also a limitation in practice that requires a lot of expertise. In this case, the initialized architectures are manually designed, which goes against the original intention of NAS. The partial fixed structure  means the architecture is partially settled. For example, in~\cite{gibb2018genetic}, a max-pooling layer is added to the network after every set of four convolution layers.

In Table~\ref{table_constrains}, the relatively few constraints category means that those methods have no restrictions like the three aspects discussed above. However, it does not imply there is no constraint. For example, in the classification task, the fully-connected layer is often used as the tail of the whole DNNs in some methods~\cite{sun2019evolving, assunccao2019automatic, wistuba2018deep}. Moreover, the maximum length is predefined in many methods including both \textit{fixed-length encoding strategy} methods~\cite{xie2017genetic} and \textit{variable-length encoding strategy} methods~\cite{sun2019evolving} resulting in preventing the method from discovering a deeper architecture. Wang \textit{et al.}~\cite{wang2018hybrid} tried to break the limit of maximum length by using a Gaussian distribution initialization mechanism. Irwin \textit{et al.}~\cite{irwin2019graph} broke the limit by using the evolutionary operators, the crossover operator and the mutation operator, to extend the depth to any size.

\begin{figure}
	\centering
	\includegraphics[width=0.8\linewidth]{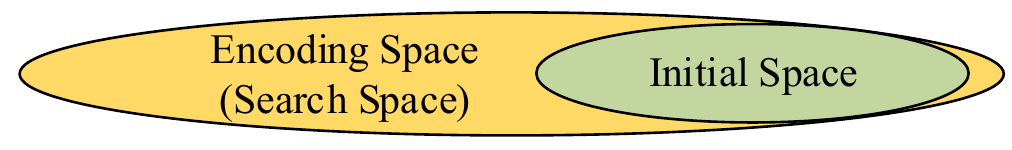}
	\caption{The relationship between encoding space, search space and initial space.}
	\label{fig_spaces}
	\vspace{-0.3cm}
\end{figure}

Generally, the encoding space can be served as the basis of the search space and the initial space. In practice, the encoding space is often the same as the search space, while the initial space is often a subspace of the encoding space. Fig.~\ref{fig_spaces} shows the relationship between the three spaces. The search space is larger than the initial space when some manual constraints are added to the population initialization. When there is no such manual constrains, search space and initial space are equivalent. Furthermore, the initial space determines what kind of individuals may appear in the initial population, and the search space determines what kind of individuals may appear in the evolutionary search process, as illustrated in Fig.~\ref{fig_flowchart}. In the following subsections, we will discuss the initial space and the search space in Subsections~\ref{sec_initial_space} and~\ref{sec_search_space}, respectively.

\subsection{Initial Space}
\label{sec_initial_space}
In general, there are three types of architecture initialization approaches in the initial space: starting from trivial initial conditions~\cite{real2017large}, randomly initialization in the encoding space~\cite{sun2019evolving} and staring from a well-designed architecture (also termed as rich initialization)~\cite{fujino2017deep}. These three types of initialization correspond to three different initial spaces: \textit{trivial space}, \textit{random space} and \textit{well-designed space}.

The \textit{trivial space} contains only a few primitive layers. For example, the LargeEvo algorithm~\cite{real2017large} initialized the population in a \textit{trivial space} where each individual constitutes just a single-layer model with no convolutions. Xie \textit{et al.}~\cite{xie2017genetic} experimentally demonstrated that a \textit{trivial space} can evolve to a competitive architecture. The reason for using as little experience as possible is to justify the advantage of EC-based methods where some novel architectures can be discovered, and most of the discovery is different from the manually designed DNN architectures. On the contrary, the \textit{well-designed space} contains the state-of-the-art architectures. In this way, a promising architecture can be obtained at the beginning of the evolution, whereas it can hardly evolve to other novel architectures. Actually, many of ENAS methods adopting this initial space focus on improving the performance upon the well-designed architecture. For example, the architecture pruning aims at compressing DNNs by removing less important connections~\cite{hu2018novel}. For a \textit{random space}, all the individuals in the initial population are randomly generated in the limited space, and it has been adopted by many methods, such as~\cite{tian2019automated,sun2019evolving,sun2019completely}. The aim of this type of initial spaces is also to reduce the intervention of human experience in the initial population.

\subsection{Search Space}
\label{sec_search_space}
After the initialization of the population, the ENAS methods start to search the architectures in the search space. Generally speaking, the search space is the same as the initial space when the random initial space is adopted. For other two types of initial spaces, however, due to the relatively small initial space, the search space will become much larger with the expectation that the promising architecture is included. It is worth noting that many methods do not directly define the search space, but restrict the search space by using evolutionary operators. For example, Irwin \textit{et al.}~\cite{irwin2019graph} did not specify the maximal depth, instead, they used the evolutionary operations to extend the architecture to any depth.

\section{Encoding Strategy}
\label{sec_architectureEncoding}
This section will discuss how to encode a network architecture into an individual of the EC methods. Each ENAS method needs to determine its encoding strategy before starting the first stage of the ENAS method, i.e., the population initialization. The most intuitive difference in different encoding strategies of ENAS methods is the length of the encoded individuals. 

Generally, the encoding strategies can be divided into two different categories according to whether the length of an individual changes or not during the evolutionary process. They are the \textit{fixed-length encoding strategy} and the \textit{variable-length encoding strategy}. Particularly, the individuals have the same length during the evolutionary process when the \textit{fixed-length encoding strategy} is used. In contrast, the individuals are with different lengths during the evolutionary process if the \textit{variable-length encoding strategy} is employed. The advantage of the \textit{fixed-length encoding strategy} is that it is easy to use standard evolutionary operations which are originally designed for the individuals with equal lengths. In Genetic CNN~\cite{xie2017genetic}, for example, the fixed-length binary string helped the evolutionary operators (especially the crossover) with easy implementation. Another example is~\cite{loni2020deepmaker},  where Loni \textit{et al.} used a fixed-length string of genes to represent the architecture which led to easy implementation of the one-point crossover in the corresponding encoded information. However, a proper maximal length must be predefined for the fixed-length encoding strategy. Because the maximal length relates to the optimal depth of the DNN architecture, which is unknown in advance, the corresponding ENAS algorithms still rely on expertise and experience. 

Compared with the \textit{fixed-length encoding strategy}, the \textit{variable-length encoding strategy} dose not require the human expertise regarding the optimal depth in advance, which has the potential to be fully automated. In addition, the advantage of this encoding strategy is that it can define more details of the architecture with freedom. For example, when solving a new task where there is no expertise in knowing the optimal depth of the DNN, we just initialize the individuals with a random depth, and the optimal depth can be found through the variable-length encoding strategy, where the depths of the corresponding DNN can be changed during the evolutionary process. However, the variable-length encoding strategy also brings some drawbacks. Because the traditional evolutionary operators might not be suitable for this kind of encoding strategy, the corresponding evolutionary operators need to be redesigned, where an example can be seen in~\cite{sun2019evolving}. Another disadvantage is that, due to the flexibility of variable length encoding, it could generally result in over-depth architectures, which sometimes further leads to more time-consuming fitness evaluation. Please note that some works claimed their use of the variable-length encoding strategy, where each individual is designed with a maximal length, and then the placeholder is used to indicate the gene's validation of the gene~\cite{wang2018evolving}. For example, the individual is designed to have $1,000$ genes, while some genes having the values of zeros do not participate in the evolutionary process. In this paper, we also categorize these methods into the fixed-length encoding strategy.

In addition, most of the DNN architectures can be represented as directed graphs which are made up of different basic units and the connections between the units. Therefore, the encoding for architecture can be divided into two aspects: configurations of basic units and connections, which will be discussed in the following subsections.

\subsection{Encoding for Configurations of Basic Units}
In practice, different basic units have different configurations, such as layers, blocks and cells, which are demonstrated in Section~\ref{sec_encoding_space}. For example, in CNNs, there are multiple parameters in the primitive layers which can be seen in Table~\ref{table_parameters}. As for the DenseBlock implemented by Sun \textit{et al.}~\cite{sun2019completely}, only two parameters are needed to build the block. Because the configuration of cell is more flexible than that of blocks in CNNs, which can be regarded as a microcosm of a complete neural network. For example, the cells in~\cite{real2019regularized} are made up by a combination of 10 layers selected from 8 different primitive layers. But the cells do not include some of the configurations of primitive layer, such as the feature map size which is an important parameter in the primitive layer~\cite{sun2019evolving}.

\subsection{Encoding for Connections}
When the parameters (configurations) of the basic units in the architecture have been determined, the corresponding architecture cannot be built up immediately. Since edges are indispensable in directed graph, connections are also part of the architecture. The connections discussed in this section include not only connections between basic units, but also the connections within the basic units.

Generally, the architectures of neural networks can be divided into two categories: \textit{linear architecture} and \textit{non-linear architecture}. The former denotes the architectures containing sequential basic units. The latter indicates that there are skip-connections or loop-connections in the architecture. Please note that, the structure could be the macro structure consisting of basic units, or the micro structure within the basic units.

\subsubsection{Linear Architecture}
The linear architecture can be found in different kinds of architectures including the one generated from the layer-based encoding space and the block-based encoding space. Its widespread use in ENAS stems from its simplicity. No matter how complex the internal of the basic units is, many ENAS methods stack the basic units one by one to build up the skeleton of the architecture which is linear. For example, in AE-CNN~\cite{sun2019completely}, Sun \textit{et al.} stacked different kinds of blocks to generate the architecture.

One special case is a linear architecture generated from the layer-based encoding space. In this case, there is no need to solely encode the connections, and only the parameters in each basic unit are enough to build an architecture. One classical example can be seen in~\cite{sun2019evolving} where Sun \textit{et al.} explored a great number of parameters based on a linear CNN architecture. However, most architectures are not designed to be linear. The skip connections in ResNet~\cite{he2016deep} and the dense connections in DenseNet~\cite{huang2017densely} show the ability to build a good architecture.

\subsubsection{Non-linear Architecture}
Firstly, we will introduce two approaches to encoding a non-linear architecture in this subsection. Specifically, the adjacent matrix is the most popular way to represent the connections for non-linear architectures. Genetic CNN~\cite{xie2017genetic} used a binary string to represent the connection, and the string can be transformed into a triangular matrix. In the binary string, ``1'' denotes that there is a connection between the two nodes while ``0'' denotes no connection in between. Lorenzo \textit{et al.}~\cite{lorenzo2018memetic} used a matrix to represent the skip connections, and this work revolved around the adjacent matrix. Back in the 1990s, Kitano \textit{et al.}~\cite{kitano1990designing} began to study the use of the adjacent matrix to represent network connection and explained the process from the connectivity matrix, to the bit-string genotype, to the network architecture phenotype. Another way to represent the connections is to  use an ordered pair $G = (V,E)$ with vertices $V$ and edge $E$ associated with a direction, to represent a directed acyclic graph. Irwin \textit{et al.}~\cite{irwin2019graph} also used this strategy to encode the connections.

Secondly, the non-linear architecture is a more common case in both the macro architecture and the micro architecture. AmoebaNet-A~\cite{real2019regularized}, as an example of non-linear macro architecture, stacked these two kinds of cells for several times to build up an architecture. In addition, each cell receives two inputs from the previous two cells separately, which means that the direct connection and the skip-connection are both used in this macro structure. Also in AmoebaNet, each cell is a non-linear architecture inside, which means the micro architecture is also non-linear. The non-linear architecture provides the architecture with more flexibility, which makes it more likely to build up a promising architecture than that of the linear ones.

\section{Population Updating}
\label{sec_Population_operators}

\begin{table*}[ht]
	\renewcommand\arraystretch{0.8}
	\caption{Categorization of EC and Different Types of Neural Network} 
	\label{table_EC}
	\vspace{-0.3cm}
	\begin{center}
		\begin{tabular}{|p{1cm}|p{0.5cm}|p{1.8cm}|p{4.5cm}|p{1.4cm}|p{2cm}|p{1.3cm}|p{2cm}|}
			\hline
			\multicolumn{3}{|l|}{}                                                                         & CNN & DBN & RNN & AE & Others  \\ \hline
			\multirow{12}{*}{\shortstack[l]{Single\\objective}}  & \multirow{3}{*}{EA}    & GAs                             &~\cite{real2017large, chen2018reinforced, xie2017genetic, singh2019genetic, gibb2018genetic, wang2018hybrid, ahmed2019novel, sapra2020evolutionary, chen2019auto, tian2019automated, dufourq2017automated, assunccao2019automatic, laredo2019automatic, wei2019automatic, sun2020automatically, shu2020automatically, sun2019completely, litzinger2019compute, sapra2020constrained, fujino2017deep, liu2019deep, wistuba2018deep, rapaport2019eegnas, zhu2019eena, dahal2020effective, johner2019efficient, chen2019efficient, kang2019efficient, song2019efficient, ren2019eigen, chung2019emotion, martin2018evodeep, kramer2018evolution, sun2019evolving, jones2019evolutionary, van2019evolutionary, wang2019evolving, piergiovanni2019evolving, hassanzadeh2020evou, assunccao2019fast, miahi2019genetic, mitschke2018gradient, liu2017hierarchical, prellberg2018lamarckian, ho2020neural, young2015optimizing, fujino2018recognizing, real2019regularized, saltori2019regularized, zhang2020sampled, shen2019searching, guo2019single, cetto2019size}  & ~\cite{lamos2012deep, zhang2019identify}    &   ~\cite{chung2019emotion, camero2018evolutionary, almalaq2018evolutionary, ortego2020evolutionary, elsaid2019evolving}        & ~\cite{hajewski2020evolutionary} & ~\cite{anwar2019boosting, sun2018evolving, shinozaki2015structure, behjat2019adaptive}  \\ \cline{3-8} 
			&                        & GP                                    & ~\cite{bi2019evolutionary, suganuma2017genetic, loni2018designing, suganuma2020evolution, evans2018evolutionary, bianco2020neural}    &     &    ~\cite{angeline1994evolutionary, rawal2018nodes}       & ~\cite{rodriguez2019evolving}  &  ~\cite{martens2019neural}    \\ \cline{3-8} 
			&                        & ES                                  &  ~\cite{fan2020evolutionary}   &     &  ~\cite{camero2019specialized, neshat2020evolutionary, tanaka2016automated, tanaka2016evolution, camero2018evolutionary, camero2019waste}         &  ~\cite{suganuma2018exploiting} &   ~\cite{loshchilov2016cma, shinozaki2015structure}     \\ \cline{2-8} 
			& \multirow{3}{*}{SI}    & ACO                                &  ~\cite{byla2019deepswarm}   &     &     ~\cite{elsaid2018optimizing, desell2015evolving, elsaid2019ant, elsaid2018using}      &       &     \\ \cline{3-8} 
			&                        & PSO                 &  ~\cite{junior2019particle, gao2020gpso, wang2018hybrid, wang2018evolving, fielding2018evolving, wang2019particle, lorenzo2017particle, passricha2019pso}   & ~\cite{qiang2019neural, kim2017particle}    &           & ~\cite{sun2018experimental, sun2018particle}   &    ~\cite{teng2019catalytic ,zhang2000particle, ye2017particle}     \\ \cline{2-8} 
			& \multirow{6}{*}{Other}   & Memetic                           & ~\cite{lorenzo2018memetic, evans2019genetic}    &     &           &       &      \\ \cline{3-8} 
			&                        & DE                                &  ~\cite{wang2018hybrid, dahou2019arabic}   &     &     ~\cite{peng2018effective}      &  ~\cite{saufi2018differential}     &    ~\cite{de2019evolution}   \\ \cline{3-8} 
			&     & HCA              & ~\cite{elsken2017simple, kwasigroch2020neural, kwasigroch2019deep}    &     &           &     &       \\ \cline{3-8} 
			&                        & CVOA                               &     &     &     ~\cite{martinez2020coronavirus}      &       &     \\ \cline{3-8} 
			&                        & Hyper-heuristic                   &     &  ~\cite{sabar2019evolutionary}   &           &          &   \\ \cline{3-8} 
			&                        & FA                       &  ~\cite{sharaf2020automated}   &     &           &        &   \\ \cline{3-8} 
			&                        & AIS        &  ~\cite{frachon2019immunecs}   &     &           &   &     \\ \hline
			\multirow{2}{*}{\shortstack[l]{Multi-\\objective}} & \multicolumn{2}{l|}{EA}                  &  ~\cite{elsken2018efficient, zhang2019new, baldeon2019adaresu, yang2019cars, huang2020deep, loni2020deepmaker, chu2019fast, lu2019multi, chu2019multi, lu2019nsga, zhu2020real, dong2018ppp, calisto2020self, hu2018novel, loni2020deepmaker, schorn2019automated}   &     ~\cite{hossain2018multiobjective, zhang2016multiobjective}    & ~\cite{bayer2009evolving}  &  &     ~\cite{roy2020development, liang2019evolutionary}      \\ \cline{2-8} 
			& \multicolumn{2}{l|}{SI}                                &  ~\cite{jiang2020efficient, wu2019multi, wang2019evolving1}   &  ~\cite{li2019day}   &           &       &     \\ \hline
		\end{tabular}
	\end{center}
\vspace{-0.5cm}
\end{table*}

This section discusses the population updating process as shown in Fig.~\ref{fig_flowchart}. Generally, the population updating varies greatly among existing ENAS algorithms because they may employ different EC methods which are with different updating mechanisms. Table~\ref{table_EC} shows the ENAS algorithms which are classified according to the EC methods that they employ and different types of DNNs that they target at. Obviously, the EA-based ENAS algorithms dominate the ENAS. To be more specific, the GA-based ENAS is the most popular approach which largely owes to the convenience of architecture representation in GA. As a result, we give a detailed introduction to the EA-based ENAS including the selection strategy at first. Immediately after, we present a summary of the SI-based ENAS methods and others separately. The corresponding multi-objective ENAS algorithms are also introduced at the end of each subsection.

\subsection{EAs for ENAS}
\begin{table}[]
	\renewcommand\arraystretch{0.8}
	\caption{Selection Strategy} 
	\label{table_selection_strategy}
	\vspace{-0.3cm}
	\begin{center}
		\begin{tabular}{p{2cm}|p{5cm}}
			\hline
			Elitism & ~\cite{lorenzo2018memetic, elsken2017simple, frachon2019immunecs, suganuma2017genetic, wu2019multi, camero2019specialized, hajewski2020evolutionary, kwasigroch2020neural, tian2019automated, sapra2020constrained, kwasigroch2019deep, kang2019efficient, suganuma2020evolution, suganuma2018exploiting, martens2019neural} \\
			\hline
			Discard the worst or the oldest & ~\cite{sapra2020evolutionary, elsaid2019evolving, saltori2019regularized, rapaport2019eegnas, zhu2019eena, real2019regularized, an2019stylenas}\\
			\hline
			Roulette & ~\cite{xie2017genetic, singh2019genetic, gibb2018genetic, ahmed2019novel, hu2018novel, chen2019auto, loni2020deepmaker, song2019efficient, hassanzadeh2020evou} \\
			\hline
			Tournament selection & ~\cite{real2017large, chen2018reinforced, bi2019evolutionary,  dufourq2017automated, laredo2019automatic, sun2020automatically, sun2019completely, wistuba2018deep, almalaq2018evolutionary, sun2019evolving, chu2019fast, miahi2019genetic, liu2017hierarchical, chen2018joint, an2019stylenas}\\
			\hline
			Others & ~\cite{elsken2018efficient, johner2019efficient} \\
			\hline
		\end{tabular}
	\end{center}
\vspace{-0.5cm}	
\end{table}

The dash box in Fig.~\ref{fig_flowchart} shows the general flow of population updating in EA-based ENAS. In this section, we will introduce the selection strategy and the evolutionary operations which collectively update the population. Specifically, the first stage of population updating is selection. The selection strategies can be classified into several types, where Table~\ref{table_selection_strategy} shows the main kinds of selection strategies. Note that the selection strategy can be not only used in choosing individuals as parents to generate offspring with the evolutionary operators, but also used in the environmental selection stage which chooses individuals to make up the next population. Zhang \textit{et al.}~\cite{zhang2020sampled} termed these two selections as mate selection and environmental selection separately.

Existing selection strategies can be divided into five categories: elitism, discarding the worst, roulette wheel selection, tournament selection and others. The simplest strategy is elitism which retains the individuals with higher fitness. However, this can cause a loss of diversity in the population, which may lead the population falling into local optima. Discarding the worst is similar to elitism, which removes the individuals with poor fitness values from the population. Real \textit{et al.}~\cite{real2019regularized} used the aging evolution which discards the oldest individual in the population. Aging evolution can explore the search space more, instead of zooming in on good models too early, as non-aging evolution would. The same selection strategy was also used in~\cite{zhang2019identify}. Zhu \textit{et al.}~\cite{zhu2019eena} combined these two approaches to discarding the worst individual and the oldest individual at the same time. Roulette wheel selection gives every individual a probability according to its fitness among the population to survive (or be discarded), regardless it is the best or not. Tournament selection selects the best one from an equally likely sampling of individuals. Furthermore, Johner \textit{et al.}~\cite{johner2019efficient} used a ranking function to choose individuals by rank.  A selection trick termed as niching was used in~\cite{prellberg2018lamarckian, kramer2018evolution} to avoid stacking into local optima. This trick allows offspring worse than parent to survive for several generations until evolving to a better one.

Most of the methods focus on preserving the well-performed individuals, however, Liu \textit{et al.}~\cite{liu2019deep} emphasized the genes more than the survived individuals, where a gene can represent any components in the architecture. They believed the individuals which consist of the fine-gene set are more likely to have promising performance.

Some selection methods aim at preserving the diversity of the population. Elsken \textit{et al.}~\cite{elsken2018efficient} selected individuals in inverse proportion to their density. Javaheripi \textit{et al.}~\cite{javaheripi2020genecai} chose the parents based on the distance (difference) during the mate selection. They chose two individuals with the highest distance to promote the exploration search.

In terms of evolutionary operations, mutation and crossover are two of the most commonly used operations in EA-based ENAS algorithms. Particularly, the mutation is only performed on a single individual, while the crossover takes two individuals to generate offspring.

The mutation operator aims to search the global optimum around the individual. A simple idea is to allow the encoded information to vary from a given range. Sun \textit{et al.}~\cite{sun2019evolving} used the polynomial mutation~\cite{deb2001multi} on the parameters of layers which are expressed by real numbers.
To make mutation not random, Lorenzo \textit{et al.}~\cite{lorenzo2018memetic} proposed a novel Gaussian mutation based on a Gaussian regression to guide the mutation, i.e., the Gaussian regression can predict which architecture may be good, and the newly generated individuals are sampled in the regions of the search space, where the fitness values are likely to be high. This makes mutation to have a ``direction".
Moreover, Maziarz \textit{et al.}~\cite{maziarz2018evolutionary} used an RNN to guide the mutation operation. In this work, the mutation operations were not sampled at random among the possible architectural choices, but were sampled from distributions inferred by an RNN. Using an RNN to control the mutation operation can also be seen in other methods such as~\cite{chu2019fast}.
Some researches investigated the diversity of the population after mutation. Qiang \textit{et al.}~\cite{qiang2019neural} used a variable mutation probability. They used a higher probability in the early stage for better exploration and a lower probability in the later stage for better exploitation. It has been effectively applied to many other methods~\cite{piergiovanni2019evolving}. To maintain the diversity of the population after the mutation operation, Tian \textit{et al.}~\cite{tian2019automated} used \textit{force mutation} and distance calculation, which ensures the individual in the population is not particularly similar to other individuals (especially the best one). Kramer \textit{et al.}~\cite{kramer2018evolution} used the (1+1)-evolutionary strategy that generates an offspring based on a single parent with bit-flip mutation, and used a mutation rate to control and niching to overcome local optima.

The intensive computational cost of ENAS presents a bottleneck which will be discussed in later sections. To reduce the unaffordable computational cost and time, some kinds of the mutation have also been designed. Zhang \textit{et al.}~\cite{zhang2020sampled} proposed an exchange mutation which exchanges the position of two genes of the individual, i.e., exchanging the order of layers. This will not bring new layers and the weights in neural networks can be completely preserved, which means that the offspring do not have to be trained from scratch. 
Chen \textit{et al.}~\cite{chen2015net2net} introduced two function-preserving operators for DNNs, and these operators are termed as network morphisms~\cite{wei2016network}. The network morphisms aim to change the DNN architecture without the loss of the acquired experience. The network morphisms change the architecture from $F(\cdot)$ to $G(\cdot)$, which satisfies the condition formulated by Equation~(\ref{equ_function-preserving}):

\begin{equation}
\label{equ_function-preserving}
\forall x,\quad F(x)=G(x)
\end{equation}
where $x$ denotes the input of the DNN. The network morphisms can be regarded as a function-preserving mutation operation. With this operation, the mutated individuals will not have worse performance than their parents. To be more specific, Chen \textit{et al.}~\cite{chen2015net2net} proposed \textit{net2widernet} to obtain a wider net and \textit{net2deepernet} to obtain a deeper net. Elsken \textit{et al.}~\cite{elsken2017simple} extended the network morphisms with two popular network operations: skip connections and batch normalization. Zhu \textit{et al.}~\cite{zhu2019eena} proposed five well-designed function-preserving mutation operations to guide the evolutionary process by the information which have already learned. To avoid local optimal, Chen \textit{et al.}~\cite{chen2019efficient} added noises in some function-preserving mutation, and in the experiment, they found that by adding noises to pure network morphism, instead of compromising the efficiency, it by contrast, improved the final classification accuracy.
Please note that all the network morphisms can only increase the capacity of a network because if one would decrease the network’s capacity, the function-preserving property could not be guaranteed~\cite{elsken2018efficient}. As a result, the architecture generated by network morphisms is only going to get larger and deeper, which is not suitable for a device with limited computing resources, such as a mobile phone. In order for the network architecture to be reduced, Elsken \textit{et al.}~\cite{elsken2018efficient} proposed the approximate network morphism, which satisfies Equation~(\ref{equ_approximate_network_morphism}):

\begin{equation}
\label{equ_approximate_network_morphism}
\forall x,\quad F(x)\approx G(x)
\end{equation}

For the crossover operator, the single-point crossover~\cite{mitchell1998introduction} is the most popular method in EA-based ENAS~\cite{gibb2018genetic, singh2019genetic, ahmed2019novel, hu2018novel} because of its implementation simplicity. However, single-point crossover can typically apply to two individuals with equal lengths only. Therefore, this cannot be applied to variable-length individuals. To this end, Sun \textit{et al.}~\cite{sun2020automatically} proposed an efficient crossover operator for individuals with variable lengths. Sapra \textit{et al.}~\cite{sapra2020evolutionary} proposed a disruptive crossover swapping the whole cluster (a sequence of layers) between both individuals at the corresponding positions rather than only focusing on the parameters of layers. Sun \textit{et al.}~\cite{sun2019evolving} used the Simulated Binary Crossover (SBX)~\cite{deb1995simulated} to do a combination of the encoded parameters from two matched layers. Please note that the encoded parameters after SBX are not the same as that of both parents, which are quite different from other crossover operators.

EAs for multi-objective ENAS is gaining more and more attention from researchers. The single objective ENAS algorithms are always concerned about only one objective, e.g., the classification accuracy, and these algorithms have only one goal: searching for the architecture with the highest accuracy. In general, most of the multi-objective ENAS algorithms aim at dealing with both the performance of the neural network and the number of parameters simultaneously~\cite{elsken2018efficient, yang2019cars, lu2019multi, lu2019nsga}.
However, these objective functions are often in conflict with each other. For example, getting a higher accuracy often requires a more complicated architecture with the need of more computational resources. On the contrary, a device with limited computational resource, e.g., a mobile phone, cannot afford such sophisticated architectures. 

The simplest way to tackle the multi-objective optimization problem is by converting it into a single objective optimization problem with weighting factors, i.e., the weighted summation method. The Equation~(\ref{equ_multi-fitness})

\begin{equation}
\label{equ_multi-fitness}
F = \lambda f_{1} + (1-\lambda) f_{2}
\end{equation}
is the classical linear form to weight two objective functions $f_{1},f_{2}$ into a single objective function, where the $\lambda \in (0, 1)$ denotes the weighting factor. In~\cite{zhang2019new,laredo2019automatic,loni2018designing,vargas2019evolving}, the multi-objective optimization problem was solved by using the available single objective optimization methods by Equation~(\ref{equ_multi-fitness}) of the weighted summation. Chen \textit{et al.}~\cite{chen2018joint} did not adopt the linear addition as the objective function, whereas using a nonlinear penalty term. However, the weights manually defined may incur bias~\cite{deb2014multi}.

Some algorithms have been designed and widely used in multi-objective optimization, such as NSGA-II~\cite{deb2002fast}, and MOEA/D~\cite{zhang2007moea}, which have been also used in ENAS methods such as~\cite{huang2020deep,loni2020deepmaker,baldeon2019adaresu}. These methods aim to find a Pareto-font set (or non-dominant set). Only these methods are in the multi-objective category of Table~\ref{table_EC}. Some researches have made improvements on these multi-objective optimization methods for better use in ENAS.
Baldeon \textit{et al.}~\cite{baldeon2019adaresu} chose the penalty based boundary intersection approach in MOEA/D because training a neural network involves nonconvex optimization and the form of Pareto Font is unknown.
LEMONADE~\cite{elsken2018efficient} divided the objective function into two categories: $f_{exp}$ and $f_{cheap}$. $f_{exp}$ denotes the expensive-to-evaluate objectives (e.g., the accuracy), while $f_{cheap}$ denotes the cheap-to-evaluate objectives (e.g., the model size). In every iteration, they sampled parent networks with respect to sparsely distribution based on the cheap objectives $f_{cheap}$ to generate offspring. Therefore, they evaluated $f_{cheap}$ more times than $f_{exp}$ to save time. Schoron \textit{et al.}~\cite{schorn2019automated} also took the use of the LEMONADE proposed by Elsken \textit{et al.}~\cite{elsken2018efficient}.
Due to that NSGA-III~\cite{deb2013evolutionary} may fall into the small model trap (this algorithm prefers small models), Yang \textit{et al.}~\cite{yang2019cars} have made some improvements to the conventional NSGA-III for favoring larger models.

\subsection{SI for ENAS}
PSO is inspired by the bird flocking or fish schooling~\cite{kennedy1995particle}, and is easy to implement compared with other SI algorithms.
Junior \textit{et al.}~\cite{junior2019particle} used their implementation of PSO to update the particles based on the layer instead of the parameters of the layer. Gao \textit{et al.}~\cite{gao2020gpso} developed a gradient-priority particle swarm optimization algorithm to handle issues including the low convergence efficiency of PSO when there are a large number of hyper-parameters to be optimized. They expected the particle to find the locally optimal solution at first, and then move to the global optimal solution.

For ACO, the individuals are generated in a quite different way. Several ants are in an ant colony\footnote{The population in ACO also termed as colony.}. Each ant moves from node to node following the pheromone instructions to build an architecture. The pheromone is updated every generation. The paths of well-performed architecture will maintain more pheromone to attract the next ant for exploitation and at the same time, the pheromone is also decaying (i.e., pheromone evaporation), which encourages other ants to explore other areas. Byla \textit{et al.}~\cite{byla2019deepswarm} let the ants choose the path from the node to node in a graph whose depth increases gradually. Elsaid \textit{et al.}~\cite{elsaid2019ant} introduced different ant agent types to act according to specific roles to serve the needs of the colony, which is inspired by the real ants species.

SI for multi-objective ENAS started only in the last two years and the research of this field is scarce which can be seen from Table~\ref{table_EC}.
Li \textit{et al.}~\cite{li2019day} used the bias-variance framework on their proposed multi-objective PSO to get a more accurate and stable architecture.
Wu \textit{et al.}~\cite{wu2019multi} used the MOPSO~\cite{coello2004handling} for neural networks pruning. The $G_{best}$ is selected according to the crowding distance in the non-dominant solutions set. Wang \textit{et al.}~\cite{wang2019evolving1} used the OMOPSO~\cite{sierra2005improving}, which selects the leaders using a crowding factor and the $G_{best}$ is selected from the leaders. To better control the balance between convergence and diversity, Jiang \textit{et. al.}~\cite{jiang2020efficient} proposed a MOPSO/D algorithm based on an adaptive penalty-based boundary intersection.

\subsection{Other EC Techniques for ENAS}
Different from GAs, the mutation of DE exploits the information from three individuals. 
Some ENAS methods like~\cite{peng2018effective,saufi2018differential} chose DE to guide the offspring generation. However, there is little difference between different DE-based ENAS algorithms.

Wang \textit{et al.}~\cite{wang2019hybrid} proposed a hybrid PSO-GA method. They used PSO to guide the evolution of the parameters in each block encoded in decimal notation. Meanwhile, using GA to guide the evolution of the shortcut connections is encoded in binary notation. Because PSO performs well on continuous optimization and GA is suitable for optimization with binary values, this hybrid method can search architectures effectively.

HCA can be interpreted as a very simple evolutionary algorithm. For example, in~\cite{elsken2017simple} the evolutionary operators only contain mutation and no crossover, and the selection strategy is relatively simple. The memetic algorithm is the hybrids of EAs and local search. Evans \textit{et al.}~\cite{evans2019genetic} integrated the local search (as gradient descent) into GP as a fine-tuning operation. The CVOA~\cite{martinez2020coronavirus} was inspired by the new respiratory virus, COVID-19. The architecture was found by simulating the virus spreads and infecting healthy individuals. Hyper-heuristic contains two levels: high-level strategy and low-level heuristics, and a domain barrier is between these two levels. Hence the high-level strategy is still useful when the application domain is changed. AIS was inspired by theories related to the mammal immune system and do not require the crossover operator compared to the GA~\cite{frachon2019immunecs}.

\section{Efficient Evaluation}
\label{sec_evaluation}
In this section, we will discuss the strategies to improve the efficiency of evaluations, with the consideration that the evaluation is often the most time-consuming stage of ENAS algorithms~\cite{zhou2021survey}.

Real \textit{et al.}~\cite{real2017large} used 250 computers to finish the LargeEvo algorithm over 11 days. Such computational resources are not available for everyone interested in NAS. Almost all of the methods evaluate individuals by training them first and evaluating them on the validation/test dataset. Since the architecture is becoming more and more complex, it will take a lot of time for training each architecture to convergence. So it is needed to investigate new methods to shorten the evaluation time and reduce the dependency on large amounts of computational resources.
Table~\ref{table_shorten_time} lists five of the most common methods to reduce the time: weight inheritance, early stopping policy, reduced training set, reduced population, and population memory. We would like to introduce the five kinds of methods first, then other kinds of promising methods next, and finally the surrogate-assisted methods at the end of this section.

Because the evolutionary operators do not completely disrupt the architecture of an individual, some parts of the newly generated individuals are the same as their parents. The weights of the same parts can be easily inherited. With the weight inheritance, the neural networks no longer need to be trained completely from scratch. This method has been used in~\cite{zhang2000particle} 20 years ago.
Moreover, as mentioned in Section~\ref{sec_Population_operators}, the network morphisms change the network architecture without loss of the acquired experience. This could be regarded as the ultimate weight inheritance because it solved the weight inheritance problem in the changed architecture part. The ultimate weight inheritance lets the new individual completely inherit the knowledge from their parents, which will save a lot of time.

\begin{figure}
	\centering
	\includegraphics[width=0.8\linewidth]{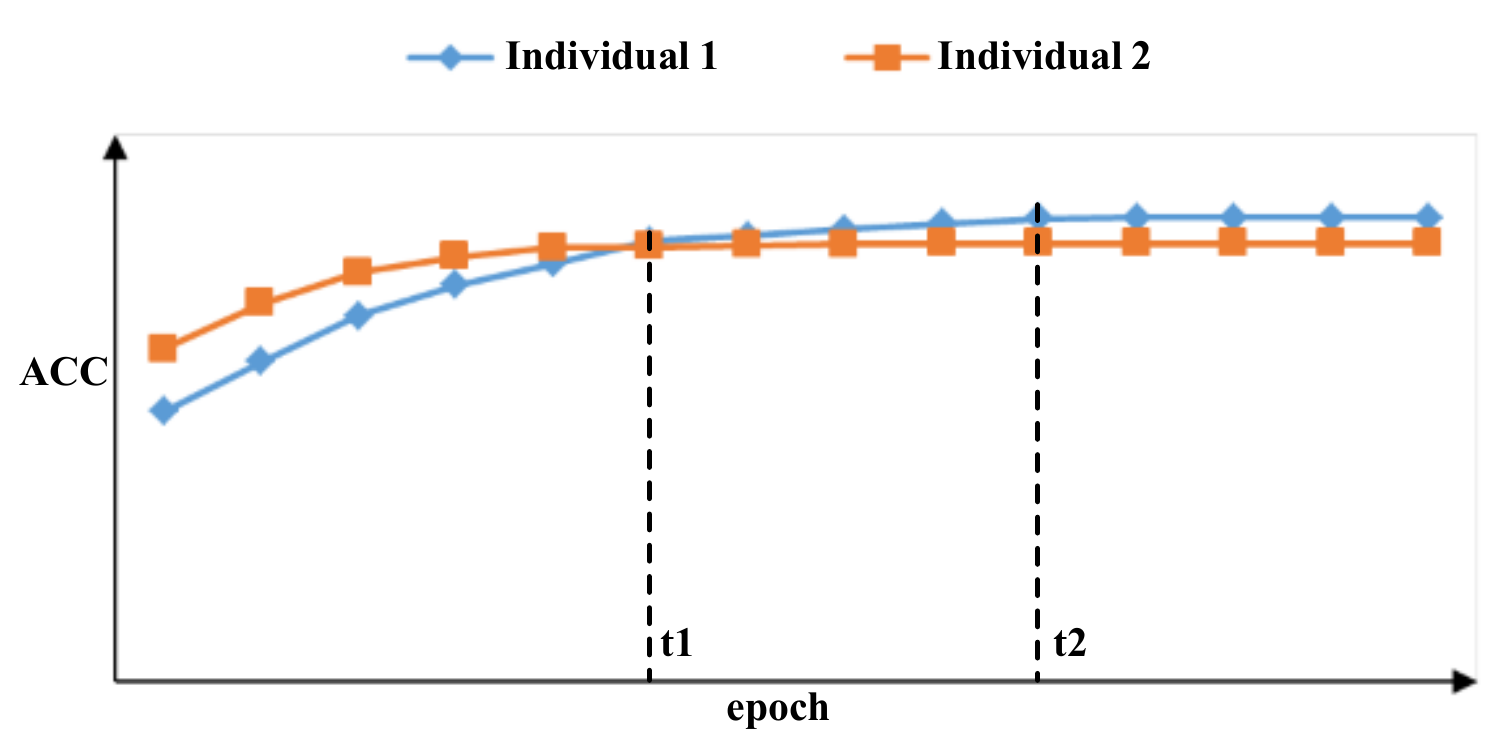}
	\caption{Two learning curve of two different individuals.}
	\label{fig_learning_curve}
	\vspace{-0.2cm}
\end{figure}

The early stopping policy is another method which has been used widely in NAS. The simplest way is to set a fixed and relatively small number of training epochs. This method is used in~\cite{sun2018particle}, where training the individuals after a small number of epochs is sufficient. Similarly, Assunccao \textit{et al.}~\cite{assunccao2019automatic} let the individuals undergo the training for the same and short time each epoch (although this time is not fixed and will increase with the epoch), to allow the promising architectures have more training time to get a more precise evaluation, So \textit{et al.}~\cite{so2019evolved} set hurdles after a fixed number of epochs. The weak individuals stop training early to save time.
However, the early stopping policy can lead to inaccurate estimation about individuals' performance (especially the large and complicated architecture), which can be seen in Fig.~\ref{fig_learning_curve}. In Fig.~\ref{fig_learning_curve}, \emph{individual2} performs better than \emph{individual1} before epoch \emph{t1}, whereas \emph{individual1} performs better in the end. Yang \textit{et al.}~\cite{yang2019cars} also discussed this phenomenon. So, it is crucial to determine at which point to stop.
Note that neural networks can converge or hardly improve its performance after several epochs, as seen in the \emph{t1} for \emph{individual2} and the \emph{t2} for \emph{individual1} in Fig.~\ref{fig_learning_curve}. Using the performance estimated at this point can evaluate an individual's relatively accurately with less training time. Therefore, some methods such as~\cite{fujino2017deep, rapaport2019eegnas} stopped training when observing there is no significant performance improvement.
Suganuma \textit{et al.}~\cite{suganuma2020evolution} used the early stopping policy based on a reference curve. If the accuracy curve of an individual is under the reference curve for successive epochs, then the training will be terminated and this individual is regarded as a poor one. After every epoch, the reference curve is updated by the accuracy curve of the best offspring. 

The reduced training set, i.e., using a subset of data that assuming similar properties to a large dataset, can also shorten the time effectively. Liu \textit{et al.}~\cite{liu2019deep} explored promising architectures by training on a subset and used transfer learning on the large original dataset. Because there are so many benchmark datasets in the image classification field, the architecture can be evaluated on a smaller dataset (e.g., CIFAR-10) first and then applied on a large dataset (such as CIFAR-100 and ImageNet~\cite{deng2009imagenet}). The smaller dataset can be regarded as the proxy for the large one.

The reduced population is a unique method of ENAS since other NAS approaches do not have a population. Assunccao \textit{et al.}~\cite{assunccao2019fast} reduced the population based on their previous algorithm~\cite{assunccao2018evolving, assunccao2019denser} to speed up the evolution. However, simply reducing the population may not explore the search space well in each epoch and may lose the global search ability. Another way is reducing the population dynamically. For instance, Fan \textit{et al.}~\cite{fan2020evolutionary} used the ($\mu+\lambda$) evolution strategy and divided the evolution into three stages with the population reduction, which aims to find the balance between the limited computing resources and the efficiency of the evolution. The large population in the first stage is to ensure the global search ability, while the small population in the last stage is to shorten the evolution time. Instead of reducing the population, Liu \textit{et al.}~\cite{liu2017structure} evaluated the downsizing architecture with smaller size at an early stage of evolution. Similarly, Wang \textit{et al.}~\cite{wang2019particle} did not evaluate the whole architecture but starting with a single block, and then the blocks were stacked to build architecture as evolution proceeds.

\begin{table}[]
	\renewcommand\arraystretch{0.8}
	\caption{Different Methods to Shorten the Evaluation Time} 
	\label{table_shorten_time}
	\vspace{-0.2cm}
	\begin{center}
		\begin{tabular}{p{2.5cm}|p{5cm}}
			\hline
			Weight inheritance & ~\cite{real2017large, chen2018reinforced, elsken2018efficient, byla2019deepswarm, lorenzo2018memetic, elsken2017simple, frachon2019immunecs, wu2019multi, ahmed2019novel, kwasigroch2020neural, schorn2019automated, sapra2020constrained, wistuba2018deep, kwasigroch2019deep, rapaport2019eegnas, zhu2019eena, dahal2020effective, chen2019efficient, fielding2018evolving, prellberg2018lamarckian} \\
			\hline
			Early stopping policy & ~\cite{frachon2019immunecs, sun2018experimental, gao2020gpso, wang2018hybrid, ahmed2019novel, sun2018particle, tian2019automated, assunccao2019automatic, laredo2019automatic, litzinger2019compute, wistuba2018deep, loni2018designing, rapaport2019eegnas, suganuma2020evolution, ortego2020evolutionary, wang2018evolving, wang2019evolving1, assunccao2019fast, passricha2019pso, fujino2017deep, rapaport2019eegnas}\\
			\hline
			Reduced training set & ~\cite{shu2020automatically, sapra2020constrained, liu2019deep, wang2019particle, liu2017structure} \\
			\hline
			Reduced population & ~\cite{liu2019deep, fan2020evolutionary, assunccao2019fast}\\
			\hline
			Population memory & ~\cite{fujino2017deep, miahi2019genetic, sun2020automatically, johner2019efficient}\\
			\hline
		\end{tabular}
	\end{center}
\vspace{-0.3cm}
\end{table}

The population memory is another category of the unique methods of ENAS. It works by reusing the corresponding architectural information that has previously appeared in the population. In the population-based methods, especially the GA based methods (e.g., in~\cite{sun2019evolving}), it is natural to maintain well-performed individuals in the population in successive generations. Sometimes, the individuals in the next population directly inherit all the architecture information of their parents without any modification, so it is not necessary to evaluate the individuals again. Fujino \textit{et al.}~\cite{fujino2017deep} used memory to record the fitness of individuals, and if the same architecture encoded in an individual appears, the fitness value is retrieved from memory instead of being reevaluated. Similarly, Miahi \textit{et al.}~\cite{miahi2019genetic} and Sun \textit{et al.}~\cite{sun2020automatically} employed a hashing method for saving pairs of architecture and fitness of each individual and reusing them when the same architecture appears again. Johner \textit{et al.}~\cite{johner2019efficient} prohibited the appearance of architectures that have appeared before offspring generation. This does reduce the time, however, the best individuals are forbidden to remain in the population which may lead the population evolve towards a bad direction.

There are many other well-performed methods to reduce the time in ENAS. Rather than training thousands of different architectures, one-shot model~\cite{bender2019understanding} trained only one SuperNet to save time. Different architectures, i.e., the SubNets, are sampled from the SuperNet with the shared parameters. Yang \textit{et al.}~\cite{yang2019cars} believed the traditional ENAS methods without using SuperNet which were less efficient for models to be optimized separately. In contrast, the one-shot model optimizes the architecture and the weights alternatively. But the weight sharing mechanism brings difficulty in accurately evaluating the architecture. Chu \textit{et al.}~\cite{chu2019fairnas} scrutinized the weight-sharing NAS with a fairness perspective and demonstrated the effectiveness. However, there remain some doubts that cannot explain clearly in the one-shot model. The weights in the SuperNet are coupled. It is unclear why inherited weights for a specific architecture are still effective~\cite{guo2019single}.

Making the use of hardware can reduce the time, too. Jiang \textit{et al.}~\cite{jiang2020efficient} used a distributed asynchronous system which contained a major computing node with 20 individual workers. Each worker is responsible for training a single block and uploading its result to the major node in every generation. Wang \textit{et al.}~\cite{wang2019evolving1} designed an infrastructure which has the ability to leverage all of the available GPU cards across multiple machines to concurrently perform the objective evaluations for a batch of individuals.
Colangelo \textit{et al.}~\cite{colangelo2019artificial, colangelo2019evolutionary} designed a reconfigurable hardware framework that fits the ENAS. As they claimed, this was the first work of conducting NAS and hardware co-optimization.

Furthermore, Lu \textit{et al.}~\cite{lu2019multi} adopted the concept of proxy models, which are small-scale versions of the intended architectures. For example, in a CNN architecture, the number of layers and the number of channels in each layer are reduced. However, the drawback of this method is obvious: the loss of prediction accuracy. Therefore, they performed an experiment to determine the smallest proxy model that can provide a reliable estimate of performance at a larger scale.

All the above methods obtain the fitness of individuals by directly evaluating the performance on the validation dataset. An alternative way is to use indirect methods, namely the performance predictors. As summarized in~\cite{sun2019surrogate}, the performance predictors can be divided into two categories: performance predictors based on the learning curve and end-to-end performance predictors, both of which are based on the training-predicting learning paradigm. This does not mean the performance predictor does not undergo the training phase at all, while it means learning from the information obtained in the training phase, and uses the knowledge learned to make a reasonable prediction for other architectures.
Ralwal \textit{et al.}~\cite{rawal2018nodes} took the use of the learning curve based predictor, where the fitness is not calculated in the last epoch but is predicted by the sequence of fitness from first epochs. Specifically, they used a Long Short Term Memory (LSTM)~\cite{hochreiter1997long} as a sequence to sequence model, predicting the final performance by using the learning results of the first several epochs on the validation dataset.
Sun \textit{et al.}~\cite{sun2019surrogate} adopted a surrogate-assisted method which is called end-to-end performance predictor. The predictor does not need any extra information about the performance of individuals to be evaluated. The performance predictor is essentially a regression model mapping from the architecture to its performance. The regression model needs to be trained with sufficient training data pairs first, and each pair consists of architecture and its corresponding performance. Specifically, they chose random forest~\cite{ho1995random} as the regression model to accelerate the fitness evaluations in ENAS. When the random forest receives a newly generated architecture as input, the adaptive combination of a huge number of regression trees which have been trained in advance in the forest gives the prediction.

\section{Applications}
\label{sec_application}
This section discussed different application fields of ENAS has involved. Generally, the ENAS algorithms can be applied to wherever DNNs can be applied. Table~\ref{table_application} shows the wide range of applications and Table~\ref{table_CIFAR} displays the performance of extraordinary ENAS methods on two popular and challenging datasets for image classification tasks, namely CIFAR-10 and CIFAR-100. Both of these two tables can show what ENAS has achieved so far.

\subsection{Overview}
Table~\ref{table_application} shows the applications of existing ENAS algorithms, which contains a wide range of real-world applications. 

\begin{table}[]
	\renewcommand\arraystretch{0.8}
	\caption{Applications of Existing ENAS algorithms.} 
	\label{table_application}
	\begin{tabular}{|p{1cm}|p{3.5cm}|p{3cm}|}
		\hline
		Category            & Applications                               & References               \\ \hline
		(1)                   & Image classification                       & ~\cite{real2017large, chen2018reinforced, elsken2018efficient, byla2019deepswarm, junior2019particle, lorenzo2018memetic, wang2018hybrid, frachon2019immunecs, sun2018experimental, xie2017genetic, suganuma2017genetic, singh2019genetic, irwin2019graph, wang2018hybrid, wu2019multi, zhang2019new, ahmed2019novel, hu2018novel, sun2018particle, sharaf2020automated, hajewski2020evolutionary, chen2019auto, schorn2019automated, tian2019automated, laredo2019automatic, sun2020automatically, yang2019cars, loshchilov2016cma, sun2019completely, litzinger2019compute, sapra2020constrained, fujino2017deep, wistuba2018deep, huang2020deep, loni2020deepmaker, loni2018designing, zhu2019eena, dahal2020effective, chen2019efficient, jiang2020efficient, kang2019efficient, ren2019eigen, martin2018evodeep, kramer2018evolution, suganuma2020evolution, colangelo2019evolutionary, evans2018evolutionary, maziarz2018evolutionary, rodriguez2019evolving, wang2018evolving, wang2019evolving, tirumala2020evolving, miikkulainen2019evolving, fielding2018evolving, sun2019evolving, wang2019evolving1, assunccao2019fast, assunccao2019fast1, javaheripi2020genecai, evans2019genetic, mitschke2018gradient, liu2017hierarchical, cheung2011hybrid, chen2020immunetnas, assunccao2020incremental, chen2018joint, prellberg2018lamarckian, lu2019multi, lu2019nsga, wei2020npenas, young2015optimizing, wang2019particle, lorenzo2017particle, real2019regularized, saltori2019regularized, zhang2020sampled, shen2019searching, dong2018ppp, guo2019single, cetto2019size, johner2019efficient}                         \\ \hline
		(1)                   & Image to image                             & ~\cite{shu2020automatically, song2019efficient, chu2019fast, chu2019multi, van2019evolutionary, suganuma2018exploiting, ho2020neural}                         \\ \hline
		(1)                   & Emotion recognition                        & ~\cite{gao2020gpso, chung2019emotion}                         \\ \hline
		(1)                   & Speech recognition                         & ~\cite{tanaka2016automated, anwar2019boosting, tanaka2016evolution, passricha2019pso, shinozaki2015structure}                         \\ \hline
		(1)                   & Language modeling                          & ~\cite{miikkulainen2019evolving, rawal2018nodes}                         \\ \hline
		(1)                   & Face De-identification                     & ~\cite{song2019learning}         \\ \hline
		(2)                   & Medical image segmentation                 & ~\cite{baldeon2019adaresu, lorenzo2018memetic, calisto2020self, zhang2019identify, fan2020evolutionary, hassanzadeh2020evou, qiang2019neural}                         \\ \hline
		(2)                   & Malignant melanoma detection               & ~\cite{kwasigroch2020neural, kwasigroch2019deep}         \\ \hline
		(2)                   & Sleep heart study                          & ~\cite{de2019evolution}         \\ \hline
		(2)                   & Assessment of human sperm                  & ~\cite{miahi2019genetic}         \\ \hline
		(3)                   & Wind speed prediction                      & ~\cite{neshat2020evolutionary}   \\ \hline
		(3)                   & Electricity demand time series forecasting & ~\cite{martinez2020coronavirus}  \\ \hline
		(3)                   & Traffic flow forecasting                   & ~\cite{li2019day}                \\ \hline
		(3)                   & Electricity price forecasting              & ~\cite{peng2018effective}        \\ \hline
		(3)                   & Car park occupancy prediction              & ~\cite{camero2018evolutionary}   \\ \hline
		(3)                   & Energy consumption prediction              & ~\cite{almalaq2018evolutionary}  \\ \hline
		(3)                   & Time series data prediction                & ~\cite{elsaid2019evolving}       \\ \hline
		(3)                   & Financial prediction                       & ~\cite{zhang2015genetic}         \\ \hline
		(3)                   & Usable life prediction                     & ~\cite{zhang2016multiobjective}  \\ \hline
		(3)                   & Municipal waste forecasting                & ~\cite{camero2019specialized}    \\ \hline
		(4)                   & Engine vibration prediction                & ~\cite{elsaid2018optimizing, elsaid2018using}     \\ \hline
		(4)                   & UAV                                        & ~\cite{behjat2019adaptive}       \\ \hline
		(4)                   & Bearing fault diagnosis                    & ~\cite{saufi2018differential}    \\ \hline
		(4)                   & Predicting general aviation flight data    & ~\cite{desell2015evolving}       \\ \hline
		(5)                   & Crack detection of concrete                & ~\cite{gibb2018genetic}          \\ \hline
		(5)                   & Gamma-ray detection                        & ~\cite{assunccao2019automatic}   \\ \hline
		(5)                   & Multitask learning                         & ~\cite{liang2018evolutionary}    \\ \hline
		(5)                   & Identify Galaxies                          & ~\cite{jones2019evolutionary}    \\ \hline
		(5)                   & Video understanding                        & ~\cite{piergiovanni2019evolving} \\ \hline
		(5)                   & Comics understanding                       & ~\cite{fujino2018recognizing, fujino2017evolutionary} \\ \hline
		
	\end{tabular}
\vspace{-0.5cm}
\end{table}

Generally, these applications can be grouped into the following five different categories:

(1) Image and signal processing, including image classification which is the most popular and competitive field, image to image processing (including image restoration, image denoising, super-resolution and image inpainting), emotion recognition, speech recognition, language modelling, and face de-identification.

(2) Biological and biomedical tasks, including medical image segmentation, malignant melanoma detection, sleep heart study, and assessment of human sperm.

(3) Predictions and forecasting about all sorts of data, including the prediction of wind speed, car park occupancy, time-series data, financial and usable life, the forecasting of electricity demand time series, traffic flow, electricity price, and municipal waste.

(4) Engineering, including engine vibration prediction, Unmanned Aerial Vehicle (UAV), bearing fault diagnosis and predicting general aviation flight data.

(5) Others, including crack detection of concrete, gamma-ray detection, multitask learning, identify galaxies, video understanding, and comics understanding.

\subsection{Comparisons on CIFAR-10, CIFAR-100 and ImageNet}
\label{sec_CIFAR}

\begin{table*}[!htbp]
	\renewcommand\arraystretch{0.8}
	\caption{The Comparison of the Classification Error Rate on CIFAR-10, CIFAR-100 and ImageNet}
	\label{table_CIFAR}
	\vspace{-0.2cm}
	\begin{tabular}{|p{3.7cm}|p{1.3cm}|p{1.8cm}|p{2.4cm}|p{2.4cm}|p{3cm}|p{0.5cm}|}
		\hline
		\textbf{ENAS Methods}                                     & \textbf{GPU Days}               & \textbf{Parameters(M)}           & \textbf{CIFAR-10(\%)}                 & \textbf{CIFAR-100(\%)}           &  {\textbf{ImageNet(Top1/Top5 \%)}} & \textbf{Year}                                                                                                                                                                                                      \\ \hline
		CGP-DCNN~\cite{ loni2018designing}                               & ---                & 1.1                      & 8.1                                 &   ---                       & --- & 2018                                                                                                                                                     \\ \hline
		EPT~\cite{ sapra2020constrained}                                 & 2                      & ---                  & 7.5                                 &    ---                           &   ---  & 2020                                                                                                                                                                                                   \\ \hline
		\multirow{2}{*}{GeNet~\cite{ xie2017genetic}}                    & \multirow{3}{*}{17}    & ---                  & 7.1                                 &       ---                    & ---  & \multirow{3}{*}{2017}                                                                                                          \\ \cline{3-6}
		&                        & ---                  &      ---                               & 29.03                               &  ---          &                                                                                                                                                                                      \\ \cline{3-6}
		& & 156 & --- & --- & {27.87 / 9.74} &
		\\ \hline
		EANN-Net~\cite{ chen2019auto}                                    & ---                & ---                  & 7.05 $\pm$ 0.02        &    ---                 &     ---       & 2019           \\ \hline
		\multirow{2}{*}{DeepMaker~\cite{loni2020deepmaker}}            & \multirow{2}{*}{3.125} & 1                        & 6.9                                 &     ---                  &   ---     &\multirow{2}{*}{2020}           \\ \cline{3-6}
		&                        & 1.89                     &    ---                                 & 24.87                               &   ---           &       \\ \hline
		\multirow{2}{*}{GeneCai (ResNet-50)~\cite{ javaheripi2020genecai} }               & \multirow{2}{*}{0.024}                  & ---                  & 6.4                                 &    ---                    &  ---  & \multirow{2}{*}{2020}                 \\ \cline{3-6}
		& & --- & --- & --- & {25.7 / 7.9} & \\ \hline
		CGP-CNN (ConvSet)~\cite{ suganuma2017genetic}                    & ---                & 1.52                     & 6.75                                &   ---          &  --- &2017                   \\ \hline
		CGP-CNN (ResSet)~\cite{ suganuma2017genetic}                     & ---                & 1.68                     & 5.98                                &  ---           &  ---    &2017                                                                                             \\ \hline
		MOPSO/D-Net~\cite{jiang2020efficient}                          & 0.33                   & 8.1                      & 5.88                                &  ---  &    ---      &2019                                                                                                                                                                                                          \\ \hline
		ReseNet-50 (20\% pruned)~\cite{ hu2018novel}                     & ---                & 6.44  & 5.85                                &      ---                       &  --- &2018                                                                                                                                                           \\ \hline
		ImmuNeCS~\cite{ frachon2019immunecs}                             & 14                     & ---                  & 5.58                                &    ---  &    --- &2019                                                                                                                                                                                                       \\ \hline
		\multirow{2}{*}{EIGEN~\cite{ ren2019eigen}}                      & 2                      & 2.6                      & 5.4                                 &            ---                     &  ---  &  \multirow{2}{*}{2018}                                                                                                                                                                                \\ \cline{2-6}
		& 5                      & 11.8                     &             ---                       & 21.9                                &     --- &                                                                                                                                                                                            \\ \hline
		\multirow{2}{*}{LargeEvo~\cite{ real2017large}}     & \multirow{2}{*}{2750}  & 5.4                      & 5.4                                 &          ---                        & ---  &    \multirow{2}{*}{2017}                                                                                                                                                                                      \\ \cline{3-6}
		&                        & 40.4                     &      ---                               & 23                                  &        ---           &        \\ \hline
		\multirow{2}{*}{CGP-CNN (ConvSet)~\cite{ suganuma2020evolution}} & 31                     & 1.5                      & 5.92 (6.48 $\pm$ 0.48) &   ---            &  ---  & \multirow{4}{*}{2019}                                                     \\ \cline{2-6}
		& ---                & 2.01                     &        ---                             & 26.7 (28.1 $\pm$ 0.83)   &  --- &                                                                                                                                                                                                  \\ \cline{1-6}
		\multirow{2}{*}{CGP-CNN (ResSet)~\cite{ suganuma2020evolution}}  & 30                     & 2.01                     & 5.01 (6.10 $\pm$ 0.89) &   ---    &             ---                     &               \\ \cline{2-6}
		& ---                & 4.6                      &   ---                                  & 25.1 (26.8 $\pm$ 1.21)   &  --- &                                                                                                                                                                                                  \\ \hline
		MOCNN~\cite{ wang2019evolving1}                                   & 24                     & ---                  & 4.49                                &  ---      &   ---  & 2019                                                                                                                                                                                                          \\ \hline
		\multirow{2}{*}{NASH~\cite{ elsken2017simple}}                   & 4                      & 88                       & 4.4                                 &        ---                        &   ---  &   \multirow{2}{*}{2017}                                                                                                                                                                                 \\ \cline{2-6}
		& 5                      & 111.5                    &     ---                               & 19.6                                &  ---   &                                                                                                                                                                                             \\ \hline
		HGAPSO~\cite{ wang2018hybrid}                                    & 7+                     & ---                  & 4.37                                &       ---                &    ---   & 2018                                                                                                                                                        \\ \hline
		\multirow{3}{*}{DPP-Net~\cite{dong2018dpp},~\cite{cheng2018searching}}                   & \multirow{3}{*}{2}                      & 11.39                       & 4.36                                 &     ---                     &   ---    &  \multirow{3}{*}{2018}         \\ \cline{3-6}
		&                       & 0.45                    &   5.84                                  &      ---                    &    ---   &                                                                                                                                                                                                 \\ \cline{3-6}
		& & 4.8 & --- & --- & {26.0 / 8.2} & \\ \hline
		\multirow{2}{*}{AE-CNN~\cite{ sun2019completely}}                & 27                     & 2                        & 4.3                                 &        ---                    &    ---     &\multirow{2}{*}{2019}                                                                                                                                                                                 \\ \cline{2-6}
		& 36                     & 5.4                      &       ---                              & 20.85                               &   ---   &                                                                                                                                                                                            \\ \hline
		SI-ENAS~\cite{zhang2020sampled}  & 1.8 & --- & 4.07 & 18.64 & --- & 2020   \\ \hline
		EPSOCNN~\cite{wang2019particle}  & 4-  & 6.77 & 3.69 & --- & --- & 2019  \\ \hline
		\multirow{2}{*}{Hierarchical Evolution~\cite{ liu2017hierarchical} }              & \multirow{2}{*}{300}                    & ---                  & 3.63 $\pm$ 0.10        &  ---  & --- &  \multirow{2}{*}{2017}                                                                                                                                                                                        \\ \cline{3-6}
		& & --- & --- & --- & {20.3 / 5.2} & \\ \hline
		\multirow{2}{*}{EA-FPNN~\cite{ wistuba2018deep}}                 & 0.5                    & 5.8                      & 3.57                                &      ---                 &  ---    & \multirow{2}{*}{2018}                                                                                                                                                  \\ \cline{2-6}
		& 1                      & 7.2                      &       ---                              & 21.74                               &   ---   &                                                                                                                                                                                            \\ \hline
		\multirow{2}{*}{AmoebaNet-A~\cite{real2019regularized}}  & \multirow{2}{*}{3150}  & 3.2  & 3.34 $\pm$ 0.06 & --- & --- & \multirow{2}{*}{2018}  \\ \cline{3-6}
		&& 86.7 & --- & --- & {17.2 / 3.9} & \\ \hline
		Firefly-CNN~\cite{ sharaf2020automated}                          & ---                & 3.21                     & 3.3                                 & 22.3                 &  ---   & 2019                                                                                                                          \\ \hline
		\multirow{3}{*}{CNN-GA~\cite{ sun2020automatically}}             & 35                     & 2.9                      & 3.22                                &   ---                 &    ---    &   \multirow{3}{*}{2018}      \\ \cline{2-6}
		& 40                     & 4.1                      &      ---                               & 20.53                               &      --- &                                                                                                                                                                                           \\ \cline{2-6}
		& 35 & --- & --- & --- & {25.2 / 7.7} & \\ \hline
		\multirow{3}{*}{JASQNet~\cite{chen2018joint}}                   & \multirow{3}{*}{3}                      & 3.3                      & 2.9                                 &     ---                       & ---   & \multirow{3}{*}{2018}           \\ \cline{3-6}
		&                       & 1.8                      & 2.97                                &               ---                      &   ---    &                                                                                                                                                                                           \\ \cline{3-6} 
		& & 4.9 & --- & --- & {27.2 / ---} & \\ \hline
		\multirow{2}{*}{RENASNet~\cite{chen2018reinforced}}                             & \multirow{2}{*}{6}                      & 3.5                      & 2.88 $\pm$ 0.02        &    ---   & --- & \multirow{2}{*}{2018}                                                                                                                  \\ \cline{3-6}
		& & 5.36 & --- & --- & {24.3 / 7.4} & \\ \hline
		NSGA-Net~\cite{ lu2019nsga}                                      & 4                      & 3.3                      & 2.75                                & ---                     &   ---  & 2018                                                                     \\ \hline
		\multirow{4}{*}{CARS~\cite{ yang2019cars}}                       & \multirow{4}{*}{0.4}   & 2.4                      & 3                                   &       ---                 &  ---   & \multirow{4}{*}{2019}                                                                                                                              \\ \cline{3-6}
		&                        & 3.6                      & 2.62                                &              ---                       & ---             &                                                                                                                                                                                    \\ \cline{3-6} 
		& & 3.7 & --- & --- & {27.2 / 9.2} & \\ \cline{3-6}
		& & 5.1 & --- & --- & {24.8 / 7.5} & \\ \hline
		\multirow{3}{*}{LEMONADE~\cite{elsken2018efficient}}            & \multirow{3}{*}{80}    & 13.1                     & 2.58                                &        ---                   &  ---  & \multirow{3}{*}{2018}                                                                                                                            \\ \cline{3-6}
		&                        & 0.5                      & 4.57                                &                  ---                   &       ---                    &                                                                                                                                                                       \\ \cline{3-6}
		& & --- & --- & --- & {28.3 / 9.6} & \\ \hline
		\multirow{2}{*}{EENA~\cite{ zhu2019eena}}                        & \multirow{2}{*}{0.65}  & 8.47                     & 2.56                                &            ---            &  ---   & \multirow{2}{*}{2019}                                                                                                       \\ \cline{3-6}
		&                        & 8.49                     &           ---                          & 17.71                               &   ---     &                                                                                                                                                                                            \\ \hline
		EEDNAS-NNMM~\cite{ chen2019efficient}                            & 0.5                    & 4.7                      & 2.55                                &           ---          &     ---    & 2019                                                                                                                                                   \\ \hline
		\multirow{5}{*}{NSGANet~\cite{lu2019multi}}                    & \multirow{5}{*}{27}    & 0.2                      & 4.67                                &     ---                  &    ---    & \multirow{5}{*}{2019}                                             \\ \cline{3-6}
		&                        & 4                        & 2.02                                &         ---                            &      ---                 &                                                                                                                                                                           \\ \cline{3-6}
		&                        & 0.2                      &             ---                        & 25.17                               &    ---                 &                 \\ \cline{3-6}
		&                        & 4.1                      &       ---                              & 14.38                               &     ---    &                                                                                                                                                                                         \\ \cline{3-6}
		& & 5.0 & --- & --- & {23.8 / 7.0} & \\ \hline
	\end{tabular}
\vspace{-0.3cm}
\end{table*}

In Table~\ref{table_application}, it is obvious to see that many ENAS methods are applied to the image classification tasks. The benchmark dataset, CIFAR-10 which contains a total of ten classes, and the CIFAR-100 is the advanced dataset including a hundred classes. These two datasets have been widely used in image classification tasks, and the accuracy on these two challenging datasets can represent the ability of the architecture searched. In addition, ImageNet~\cite{deng2009imagenet} is a more challenging benchmark dataset, which contains 1,000 classes and more than one million images. Because CIFAR-10 and CIFAR-100 are relatively small and easy to be over-fitting nowadays~\cite{chen2018reinforced}, the results of the ENAS methods on ImageNet are also included. We have collected the well-performed ENAS methods tested on these three datasets and showed the results in Table~\ref{table_CIFAR}, where the methods are ranked in an ascending order of their best accuracy on CIFAR-10, i.e., the methods are ranked in a descending order of their error rate based on the conventions. The data shown under column ``CIFAR-10" and ``CIFAR-100" denotes the error rate of each method on the corresponding datasets. Especially, as for ImageNet, we report both top1 and top5 error rates. Furthermore, the ``GPU Days", which was initially proposed in~\cite{sun2019evolving}, denotes the total search time of each method, it can be calculated by Equation~(\ref{equ_GPUDays})

\begin{equation}
\label{equ_GPUDays}
GPU\ Days = The\ number\ of\ GPUs \times t
\end{equation}
where the $t$ denotes the number of days that each method searched for. In Table~\ref{table_CIFAR} ``Parameters" denotes the total number of parameters which can represent the capability of architecture and the complexity. In addition, the symbol ``---" in Table~\ref{table_CIFAR} implies there is no result publically reported by the corresponding paper. The year reported in this table is its earliest time made public. Furthermore, there are additional notes provided for Table~\ref{table_CIFAR}. Firstly, if there are several results reported from literature, such as CGP-DCNN~\cite{loni2018designing} and CARS~\cite{yang2019cars}, we choose to report one or two of the most representative architectures. Secondly, we name two algorithms without proper names in Table~\ref{table_CIFAR} (i.e., Firefly-CNN~\cite{sharaf2020automated} and EEDNAS-NNMM~\cite{chen2019efficient}) based on the EC methods they used and the first letter of the title. Thirdly, we report the classification errors of CGP-CNN~\cite{suganuma2020evolution} in the format of ``best (mean $\pm$ std)". Finally, the symbols of ``+" and ``-" in the column of ``GPU Days" denote the meaning of ``more than" and ``less than", respectively.

In principle, there is no totally fair comparison. Due to the following two reasons: (1) The encoding space including the initial space and the search space is different from each other. There are two extreme cases in the initial space: trivial initialization which starts at the simplest architecture and rich initialization which starts from a well-designed architecture (e.g., ResNet-50~\cite{he2016deep}). Meanwhile the size of the search space is largely different, e.g., Ref~\cite{singh2019genetic} only takes the kernel size into the search space. (2) Different tricks exploited in the methods, e.g., the ``cutout", can make the comparisons unfair, too. The ``cutout" refers to a regularization method~\cite{devries2017improved} used in the training of CNNs, which could improve the final performance appreciably.

Table~\ref{table_CIFAR} shows the progress of ENAS for image classification according to the accuracy on CIFAR-10: LargeEvo algorithm~\cite{real2017large} (5.4\%, 2017), LEMONADE~\cite{elsken2018efficient} (2.58\%, 2018), and NSGANet~\cite{lu2019multi} (2.02\%, 2019). Many ENAS methods have a lower error rate than ResNet-110~\cite{he2016deep} with 6.43\% error rate on CIFAR-10, which is a manually well-designed architecture. Therefore, the architecture found by ENAS can reach the same level or exceed the architecture designed by experts. It shows that the ENAS is reliable and can be used in other application fields.

\section{Challenges and Issues}
\label{sec_challenges_and_issues}
Despite the positive results of the existing ENAS methods, there are still some challenges and issues which need to be addressed. In the following, we will briefly discuss them.

\subsection{The Effectiveness}
The effectiveness of ENAS is questioned by many researchers. Wistuba \textit{et al.}~\cite{wistuba2019survey} noticed that the random search can get a well-performed architecture and has proven to be an extremely strong baseline. Yu \textit{et al.}~\cite{yu2019evaluating} showed the state-of-the-art NAS algorithms performed similarly to the random policy on average. Liashchynskyi \textit{et al.}~\cite{liashchynskyi2019grid} compared with grid search, random search, and GA for NAS, which showed that the architecture obtained by GA and random search have similar performance. There is no need to use complicated algorithms to guide the search process if random search can outperform NAS based on EC approaches.

However, the evolutionary operator in~\cite{liashchynskyi2019grid} only contains a recombination operator, which limited the performance of ENAS algorithms. Although random search can find a well-performed architecture in the experiments, it cannot guarantee that it will find a good architecture every time. Moreover, recent researches~\cite{liu2017hierarchical, guo2019single} also showed the evolutionary search was more effective than random search. Furthermore, the experiments in Amoebanet~\cite{real2019regularized} and NAS-Bench-201~\cite{dong2020bench} also showed that ENAS can search for and find better architectures. Thus, there is an urgent need to design an elaborated experiment to reveal what components in ENAS are responsible for the effectiveness of the algorithm, especially in a large encoding space.

In Section~\ref{sec_Population_operators}, two types of operators have been introduced. We note that some methods like the LargeEvo algorithm~\cite{real2017large} only use single individual based operator (mutation) to generate offspring. The main reason that they did not involve the crossover operator in their method come from two reasons: the first is for simplicity~\cite{bayer2009evolving}, and the second is that simply combining a section of one individual with a section of another individual seems ``ill-suited" to the neural network paradigm~\cite{frachon2019immunecs}. Secondly in~\cite{wistuba2019survey}, the authors believed that there was no indication that a recombination operation applied to two individuals with high fitness would result in an offspring with similar or better fitness.

However, the supplemental materials in~\cite{sun2020automatically} demonstrated the effectiveness of the crossover operator in this method. This method can find a good architecture with the help of the crossover operation. On the contrary, when crossover is not used, the architecture found is not promising, unless it runs for a long time. In fact, the mutation operator let an individual explore the neighbouring region, and it is a gradually incremental search process like searching step by step. Crossover (recombination) can generate offspring dramatically different from the parents, which is more likely a stride. So, this operator has the ability to efficiently find a promising architecture. Chu \textit{et al.}~\cite{chu2019multi} preferred that while a crossover mainly contributes to exploitation, a mutation is usually aimed to introduce exploration. These two operators play different roles in the evolutionary process. But there is not a sufficient explanation of how the crossover operator works. Maybe some additional experiments need to be done on the methods without the crossover operator. 

The EC approach generally performs well when dealing with practical problems (e.g., the vehicle routing problem). This is mainly because EC is designed collectively by considering the domain knowledge in most cases~\cite{wang2015multiobjective, zhou2014local}. Similarly, another key reason to ensure the effectiveness of the ENAS algorithm is that the EC approach can effectively consider (and embed) some domain knowledge of the neural architecture. Since the original intention of ENAS is to design neural architectures automatically without manual experience, the future development of ENAS will need to continue to reduce any artificial experience by including prior knowledge of the encoding space. In this case, how to effectively embed domain knowledge into ENAS is a big challenge.

\subsection{Scalability}
The scale of the datasets used in most ENAS methods is relatively large. Taking the image classification task as an example, the MNIST dataset~\cite{lecun1998gradient} is one of the earliest datasets. In total, it contains $70, 000$ $28\times28$ grayscale images. Later in 2009, CIFAR-10 and CIFAR-100~\cite{krizhevsky2009learning} including $60, 000$ $32\times32$ color images are medium-scale datasets. One of the most well-known large-scale datasets is ImageNet~\cite{deng2009imagenet}, which provides more than 14 million manually annotated high-resolution images. Unlike CIFAR-10 and CIFAR-100, which are commonly used in ENAS, fewer methods choose to verify their performance on ImageNet~\cite{chen2018reinforced, hu2018novel, maziarz2018evolutionary}. This can be explained by the data shown in Table~\ref{table_CIFAR}, where the GPU Days are usually tens or even thousands. It is unaffordable for most researchers when the medium-scale dataset changes to the lager one.

However, Chen \textit{et al.}~\cite{chen2018reinforced} believed that the results on the larger datasets like ImageNet are more convincing because CIFAR-10 is easy to be over-fitting. A popular way to deal with this is using a proxy on CIFAR-10 and transfers to ImageNet~\cite{elsken2018efficient, real2019regularized}. Another alternative approach is to use the down-scaled dataset of a large-scale dataset such as ImageNet$64\times64$~\cite{chrabaszcz2017downsampled}. 

\subsection{Efficient Evaluation Method and Reduce Computational Cost}
Section~\ref{sec_evaluation} has introduced the most popular and effective ways to reduce the fitness evaluation time and the computational cost. In a nutshell, it can be described as a question that strikes the balance between the time spent and the accuracy of the evaluation. Because of the unbearable time for fully training the architecture, we must compromise as little as we can on the evaluation accuracy in exchange for significant reduction in the evaluation time without sufficient computing resources.

Although a lot of ENAS methods have adopted various means to shorten the evaluation time. Even though Sun \textit{et al.}~\cite{sun2019surrogate} specifically proposed a method for acceleration, the research direction of search acceleration is just getting started. The current approaches have many limitations that need to be addressed. For example, although the LargeEvo algorithm~\cite{real2017large} used the weight inheritance to shorten the evaluation time and reduced the computational cost, it still ran for several days with the use of lots of computational resources which cannot be easily accessed by many researchers. Furthermore, there is no baseline and common assessment criteria for search acceleration methods. It is a major challenge to propose a novel method to evaluate the architecture accurately and quickly.

\subsection{Interpretability}
CNNs are known as black-box-like solutions, which are hard to interpret~\cite{bi2019evolutionary}. Although some works have been done to visualize the process of feature extraction~\cite{zeiler2014visualizing}, they are uninterpretable due to a large number of learned features~\cite{evans2018evolutionary}. The low interpretability of the manual designed architecture becomes a big obstacle to the development of neural networks. To overcome this obstacle, some researches~\cite{evans2019genetic, evans2018evolutionary, bi2019evolutionary} used GP to automatically design the neural network. Being well-known for its potential interpretability, GP aims at solving problems by automatically evolving computer programs~\cite{koza1992genetic}.

All the above researches~\cite{evans2019genetic, evans2018evolutionary, bi2019evolutionary} gave a further analysis to demonstrate the interpretability. Specifically, Evans \textit{et al.}~\cite{evans2018evolutionary} have made a visualization on the JAFFE dataset~\cite{cheng2010facial} to expound how the evolved convolution filter served as a form of edge detection, and the large presence of white color in the output of the convolution can help the classification. In their subsequent work~\cite{evans2019genetic}, they made a visualization of the automatically evolved model on the Hands dataset~\cite{triesch1996robust}, where the aggregation function extracts the minimum value of a specific area in the hand image to determine whether the hand is open or closed. Furthermore, Bi \textit{et al.}~\cite{bi2019evolutionary} displayed the features described by the evolved functions like convolution, max-pooling and addition, and the generated salient features are discriminative for face classification.

Despite the interpretability the existing work made, all these GP-based ENAS methods only aim at shallow NNs and the number of the generated features is relatively small. However, all the most successful NNs have a deep architecture. This is due partly to the fact that very few GP based methods can be run on GPUs. It is necessary to use deep-GP to evolve a deeper GP tree and make a further analysis on the deeper architecture in the future.

\subsection{Future Applications}
Table~\ref{table_application} shows various applications which have been explored by ENAS. But these are just a small part of all areas of neural network applications. ENAS can be applied wherever neural networks can be applied, and automate the process of architecture designed which should have been done by experts. Moreover, plenty of the image classification successes of ENAS have proven that ENAS has the ability to replace experts in many areas. The automated architecture design is a trend.

However, this process is not completely automated. The encoding space (search space) still needs to be designed by experts for different applications. For example, for the image processing tasks, CNNs are more suitable, so the encoding space contains the layers including convolution layers, pooling layers and fully connected layers. For the time-series data processing, RNNs are more suitable, so the encoding space may contain the cells including $\Delta$-RNN cell, LSTM~\cite{hochreiter1997long}, Gated Recurrent Unit (GRU)~\cite{chung2014empirical}, Minimally-Gated Unit (MGU)~\cite{zhou2016minimal}, and Update-Gated RNN (UGRNN)~\cite{collins2016capacity}. The two manually determined encoding spaces already contain a great deal of artificial experience and the components without guaranteed performance are excluded. The problem is: can a method search the corresponding type of neural network for multiple tasks in a large encoding space, including all the popular widely used components? Instead of searching one multitask network~\cite{liang2018evolutionary} which learns several tasks at once with the same neural network, the aim is to find appropriate networks for different tasks in one large encoding space.

\subsection{Fair Comparisons}
Section~\ref{sec_CIFAR} gives a brief introduction of the unfair comparisons. The unfairness mainly comes from two aspects: (1) the tricks including cutout~\cite{devries2017improved}, ScheduledDropPath~\cite{zoph2018learning}, etc. (2) The different encoding spaces. For aspect (1), some ENAS methods~\cite{sun2020automatically} have reported the results with and without the tricks. For aspect (2), the well-designed search space is widely used in different ENAS methods. For instance, the NASNet search space~\cite{zoph2018learning} is also used in~\cite{real2019regularized, saltori2019regularized} because it is well-constructed so that even random search can perform well. The comparison under the same condition can tell the effectiveness of different search methods.

Fortunately, the first public benchmark dataset for NAS, the NAS-Bench-101~\cite{ying2019bench} has been proposed. The dataset contains 432K unique convolutional architectures based on the cell-based encoding space. Each architecture can query the corresponding metrics, including test accuracy, training time, etc., directly in the dataset without the large-scale computation. NAS-Bench-201~\cite{dong2020bench} was proposed recently and is based on another cell-based encoding space, which does not have no limits on edges. Compared with NAS-Bench-101, which was only tested on CIFAR-10, this dataset collects the test accuracy on three different image classification datasets (CIFAR-10, CIFAR-100, ImageNet-16-120~\cite{chrabaszcz2017downsampled}). But the encoding space is relatively small, and only contains 15.6K architectures. Experiments with different ENAS methods on these benchmark datasets can get a fair comparison and it will not take too much time. However, these datasets are only based on the cell-based encoding spaces and cannot contain all the search space of the existing methods, because the other basic units (layers and blocks) are built using more hyper-parameters, which may lead to a larger encoding space.

In the future, a common platform making fair comparisons needs to be built. This platform must have several benchmarks encoding space, such as the NASNet search space, NAS-Bench-101 and NAS-Bench-201. All the ENAS methods can be directly tested on the platform. Furthermore, this platform also needs to solve the problem that different kinds of GPUs have different computing power, which may cause an inaccurate GPU Days based on different standards. The GPU Days cannot be compared directly until they have a common baseline of computing power.

\section{Conclusions}
\label{sec_conclusion}
This paper provides a comprehensive survey of ENAS. We introduced ENAS from four aspects: population representation, encoding space, population updating, and fitness evaluation following the unified flow, which can be seen in Fig.~\ref{fig_flowchart}. The various applications and the performance of the state-of-the-art methods on image classification are also summarized in tables to demonstrate the wide applicability and the promising ability. Challenges and issues are also discussed to identify future research direction in this field.

To be specific, firstly, the encoding space is introduced by categories. We divide the encoding space into two parts: initial space and search space, where the former one defines the initial conditions whereas the latter one defines the architectures that can be found in the evolutionary process. Also, different encoding strategies and architecture representations are discussed. Secondly, the process of population updating including the evolutionary operators, the multi-objective search strategy, and the selection strategy are presented. A variety of EC paradigms use respective metaphors to generate new individuals. Based on standard algorithms, many improved methods have been proposed to obtain a stable and reliable search capability. Furthermore, we introduce the existing methods to reduce the need for large amounts of time and computing resources, which is a huge obstacle to efficiency.

Although the state-of-the-art methods have achieved some success, ENAS still faces challenges and issues. The first important issue is whether the EC-based search strategy has advantages. If the result is at the same level as the baseline (e.g., random search), it is unnecessary to design the complex evolutionary operators. A sophisticated experiment is in urgent need to tell the effectiveness, especially in a large encoding space. Secondly, the crossover operator is a multi-individual based operator and there is no sufficient explanation to how the crossover operator works well on ENAS. Besides, ENAS is just beginning a new era, so there is a lot of uncharted territory to be explored. 
Moreover, a unified standard or platform is demanded to make a fair comparison.

\bibliographystyle{IEEEtran}
\bibliography{IEEEabrv,mybibfile}

\end{document}